%% file: top.tex
\documentclass[10pt,twocolumn,letterpaper]{article}

\usepackage{cvpr}
\usepackage{times}
\usepackage{epsfig}
\usepackage{graphicx}
\usepackage{amsmath}
\usepackage{amssymb}
\usepackage{subcaption}
\usepackage{rotating}
\usepackage{multirow}
\usepackage{hhline}
\usepackage{float}
\usepackage{placeins}

\usepackage[pagebackref=true,breaklinks=true,letterpaper=true,colorlinks,bookmarks=false]{hyperref}

\cvprfinalcopy 

\def\cvprPaperID{2179} 

\ifcvprfinal\fi
\begin{document}

\title{Deep Watershed Transform for Instance Segmentation}

\author{Min Bai\qquad Raquel Urtasun\\
Department of Computer Science, University of Toronto\\
{\tt\small \{mbai, urtasun\}@cs.toronto.edu}
}
\maketitle

\input{abstract}

\input{intro}

\input{related}
\input{review}

\input{method}

\input{results}
\input{conclusion}

\vspace{-3.5mm}

\paragraph{Acknowledgements: } This work was partially supported by ONR-N00014-14-1-0232, Samsung, NVIDIA, Google and NSERC.

{\small
\bibliographystyle{ieee}
\bibliography{min_cvpr}
}

\end{document}

%% file: abstract.tex

\begin{abstract}

Most contemporary approaches to instance segmentation use complex pipelines involving conditional random fields, recurrent neural networks, object proposals, or template matching schemes. In this paper, we present a simple yet powerful end-to-end  convolutional neural network to tackle this task. Our approach combines intuitions from the classical watershed transform  and modern deep learning to produce an energy map of the image where object instances are unambiguously represented as energy basins. We then perform a cut at a single energy level to directly yield connected components corresponding to object instances. Our model achieves more than double the performance over the state-of-the-art on the challenging Cityscapes Instance Level Segmentation task.

\end{abstract}

%% file: intro.tex

\section{Introduction}\label{intro}

Instance segmentation seeks to identify the semantic class of each pixel as well as  associate each pixel with a physical instance of an object. This is in contrast with  semantic segmentation, which is only concerned with the first task. Instance segmentation is particularly challenging in street scenes, where the scale of the objects  can vary tremendously. Furthermore, the appearance of objects are affected by partial occlusions, specularities, intensity saturation, and motion blur. 
Solving this task will, however,  tremendously benefit applications such as  
object manipulation in robotics, or scene understanding and tracking for self-driving cars. 

\begin{figure*}[t]
\vspace{-0.5cm}
\begin{center}

\begin{subfigure}[b]{0.245\textwidth} \centering \includegraphics[width=\textwidth]{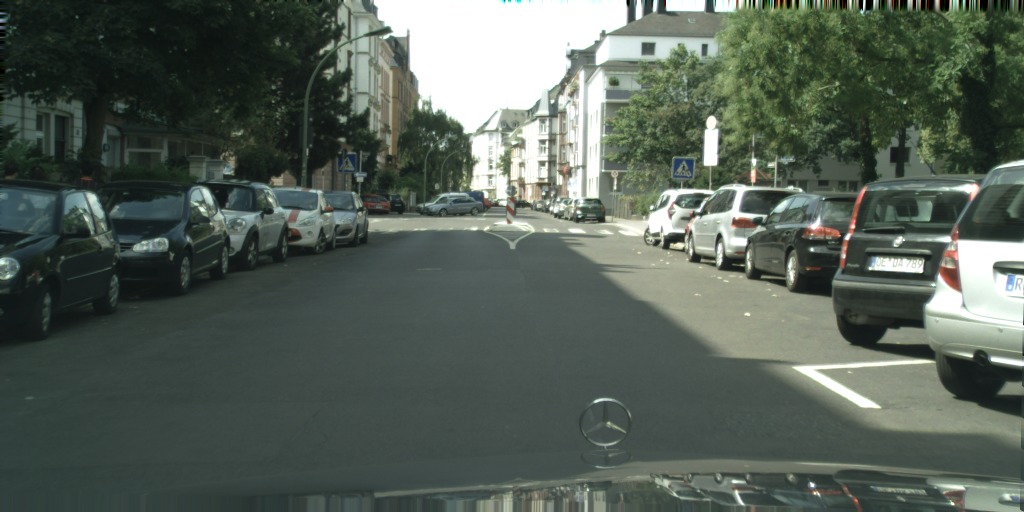}
\caption{Input Image} \end{subfigure}
\begin{subfigure}[b]{0.245\textwidth} \centering \includegraphics[width=\textwidth]{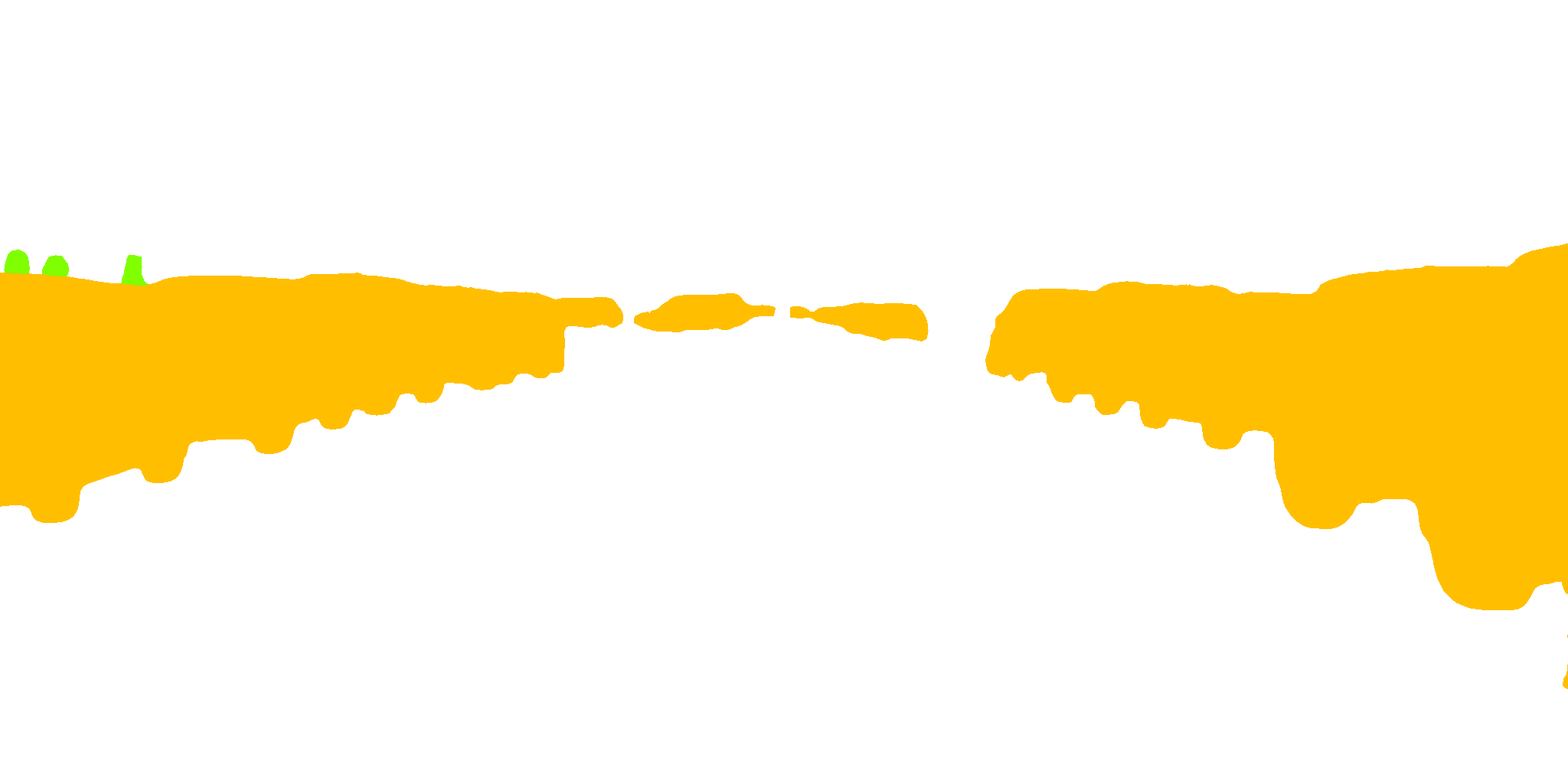}
\caption{Semantic Segmentation \cite{PSPNet}} \end{subfigure}
\begin{subfigure}[b]{0.245\textwidth} \centering \includegraphics[width=\textwidth]{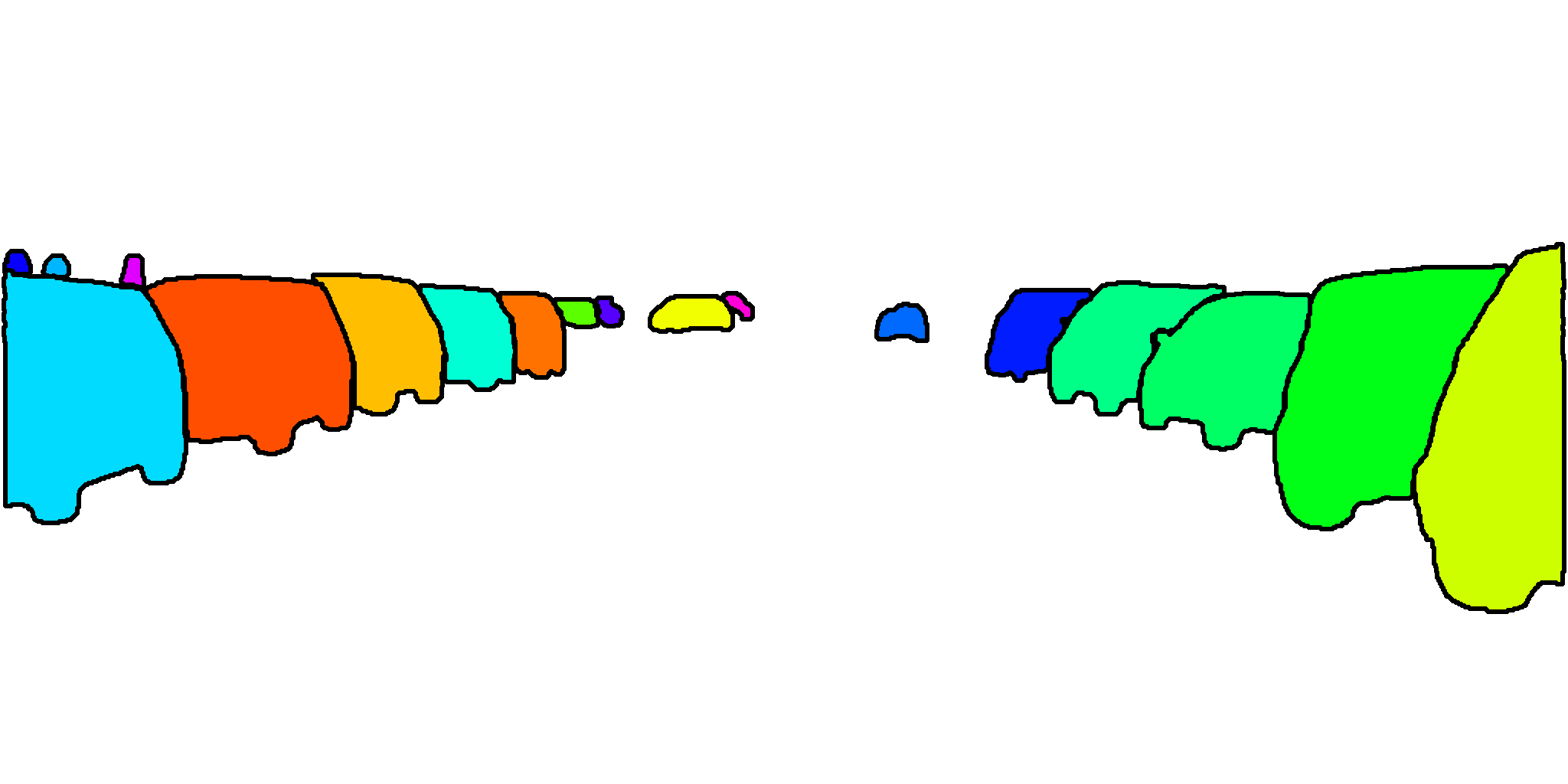} \caption{Our Instance Segmentation} \end{subfigure}
\begin{subfigure}[b]{0.245\textwidth} \centering \includegraphics[width=\textwidth]{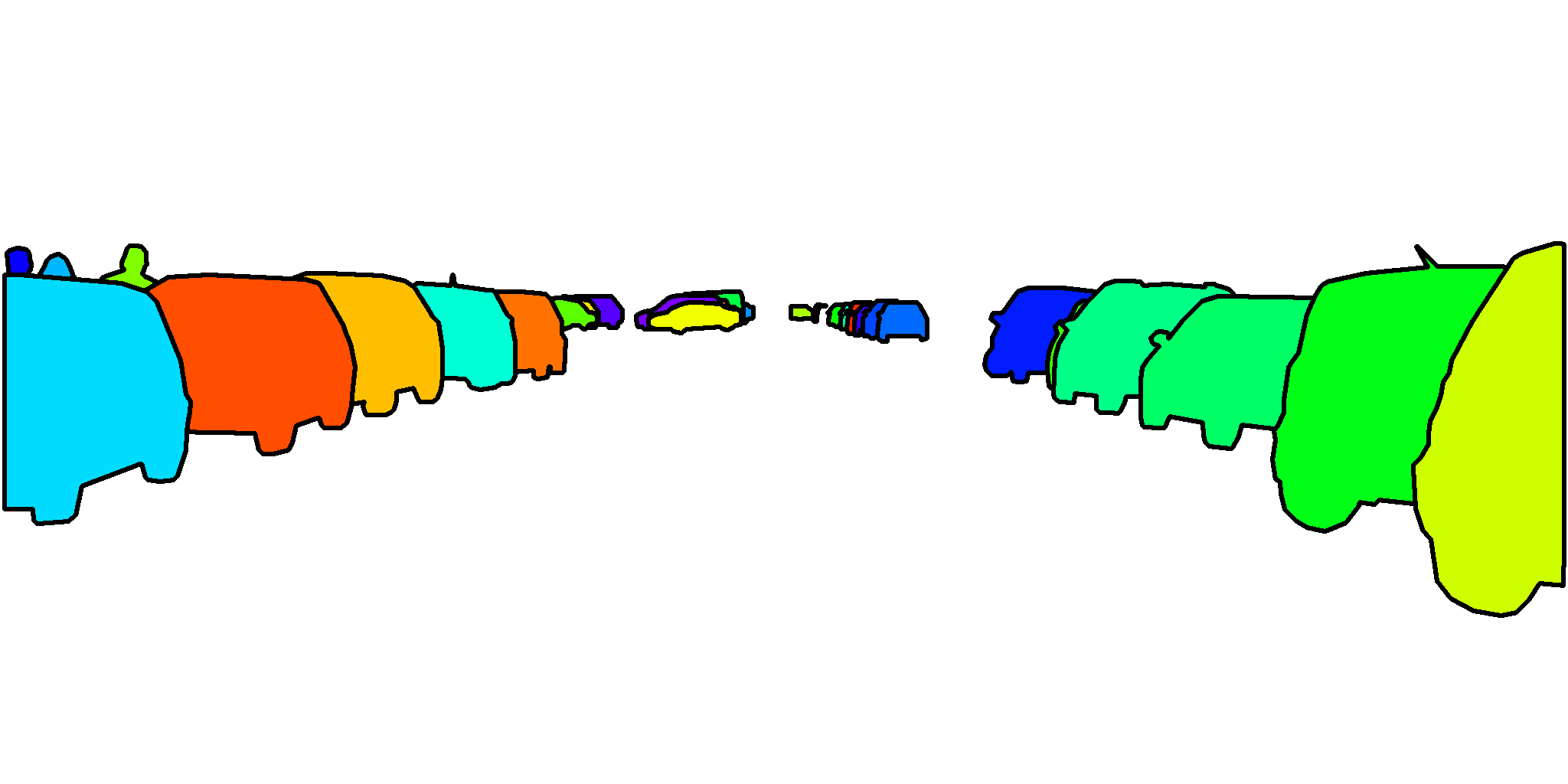} 
\caption{GT Instance Segmentation}\end{subfigure}
\vspace{-0.3cm}
\caption{Sample prediction: the input image is gated by sem. segmentation from \cite{PSPNet} and passed through our model.}\label{fig:intro_fig}
\end{center}
\end{figure*}

Current approaches generally use complex pipelines to handle instance extraction involving object proposals \cite{Pinheiro_NIPS2015_DeepMask,Pinheiro_ECCV2016_SharpMask,Dai_CVPR2016_cascades}, conditional random fields (CRF) \cite{Zhang_CVPR2016_instance,Zhang_ICCV2015_instances}, large recurrent neural networks (RNN)  \cite{Paredes_ECCV2016_RecurrentInstance,Ren_2016_endtoendinstance,Arnab_BMVC2016_CRF}, or template matching \cite{Uhrig_GCPR2016_Pixel}. In contrast, we present an exceptionally simple and intuitive method that significantly outperforms the  state-of-the-art. 
In particular, we derive a novel approach which brings together classical grouping techniques and modern deep neural networks.

The watershed transform is a well studied method in mathematical morphology. Its application to image segmentation can be traced back to the 70's \cite{Beucher_1976_watershed,Beucher_1991_watershed}. 
The idea behind this transform is fairly intuitive. 
Any greyscale image can be considered as a topographic surface. If we flood this surface from its minima and prevent the merging of the waters coming from different sources, we effectively partition the image into different components (i.e., regions). 
This transformation is typically applied to the image gradient, thus the basins correspond to  homogeneous regions in the image. 
A significant limitation of the watershed transform is its propensity to  over-segment the image. 
One of the possible solutions is to estimate the locations of object instance markers, which guide the selection of a subset of these basins  \cite{Grau_2004_Medical,Urtasun_ICIP2001_markers}.  Heuristics on the relative depth of the basins can be exploited in order to merge basins. 
However, extracting appropriate markers and creating good heuristics is difficult in practice. As a consequence, modern techniques for instance segmentation do not exploit the watershed transform. 

In this paper, we propose a novel approach which combines the strengths of modern deep neural networks with the power of this classical bottom-up grouping technique. 
We propose to directly learn the energy of the watershed transform such that each basin corresponds to a single instance, while all dividing ridges are at the same height in the energy domain. As a consequence, the components can be extracted by a cut at a single energy level without leading to over-segmentation. Our approach has several key advantages: it can be easily trained end-to-end, and produces very fast and accurate estimates. Our method does not rely on iterative strategies such as RNNs, thus has a constant runtime regardless of the number of object instances. 

We demonstrate the effectiveness of our approach in the challenging Cityscapes Instance Segmentation benchmark \cite{Cityscapes}, and show that we more than double the performance of the current state-of-the-art. 
In the following sections, we first review related work. We then present the details behind our intuition and model design, followed by an analysis of our model's performance. Finally, we explore the impact of various parts of our model in ablation studies.

%% file: related.tex
\section{Related}


Several instance level segmentation approaches have been proposed in  recent years. We now briefly review them. 

\paragraph{Proposal based:} Many approaches are based on the refinement of object proposals. For example, \cite{Arbelaez_CVPR2014_MCG} generates object segments proposals, and reasons about combining them into object instances. In a similar spirit, \cite{Hariharan_ECCV2014_SDS} selects proposals using CNN features and non-maximum suppression. Based on this, \cite{Chen_CVPR2015_occlusion} further reasons about multiple object proposals to handle occlusion scenarios where single objects are split into multiple disconnected patches. \cite{Dai_CVPR2016_cascades} uses a deep cascaded neural network to propose object instance bounding boxes, refine instance masks, and semantically label the masks in sequence. 
 \cite{Pinheiro_NIPS2015_DeepMask,Pinheiro_ECCV2016_SharpMask}  generate segmentation proposals using deep CNNs, which are then further refined to achieve better segmentation boundaries. Additionally, \cite{Zagoruyko_BMVC2016_Multipath} uses a modified R-CNN model to propose instance bounding boxes, which are then further refined to obtain instance level segmentation. 

\vspace{-4.5mm}

\paragraph{Deep structured models:}\cite{Zhang_CVPR2016_instance,Zhang_ICCV2015_instances} first use CNNs to perform local instance disambiguation and labelling, followed by a global conditional random field (CRF) to achieve instance label consistency. Recent work by \cite{Arnab_BMVC2016_CRF} uses object detection proposals in conjunction with a deep high order CRF to reason about pixel assignment in overlapping object proposal boxes. 

\vspace{-4.5mm}

\paragraph{Template matching:} \cite{Uhrig_GCPR2016_Pixel} extracts image features using CNNs to assign a sector label to each pixel in an object instance, which corresponds to one of eight discretized radial bins around the object's visible center. A template matching scheme is then used to associate instance center proposals and pixels with an object instance. 

\vspace{-4.5mm}

\paragraph{Recurrent Networks:} \cite{Paredes_ECCV2016_RecurrentInstance} uses CNNs for feature extraction, followed by a recurrent neural network (RNN) that generates instance labels for one object at a time. The recurrent structures (based on ConvLSTM \cite{Shi_NIPS2015_convlstm}) keep track of instances that have already been generated, and inhibit these regions from further instance generation. Additionally, \cite{Ren_2016_endtoendinstance} extracts image features similar to \cite{Uhrig_GCPR2016_Pixel} and employs a fairly complex pipeline including a ConvLSTM structure to direct a bounding box generation network followed by a segmentation network that extracts individual instances. 

\vspace{-4.5mm}

\paragraph{CNN:} \cite{Liang_arxiv_ProposalFree} leverages only a CNN trained to provide multiple outputs to simultaneously predict instance numbers, bounding box coordinates, and category confidence scores for each pixel. This is followed by generic clustering algorithms to group the resulting output into instance-wise labels. Additionally, \cite{Li_CVPR2016_iterative} proposed deep convolutional neural network that learns the underlying shapes of objects, and performs multiple unconstrained inference steps to refine regions corresponding to an object instance while ignoring neighboring pixels that do not belong to the primary object.  

\vspace{-4.5mm}

\paragraph{Proposal + recursion: } \cite{Liang_CVPR2016_reversible} proposed a novel method that recursively refines proposals. 

\vspace{-4.5mm}

\paragraph{} In contrast, in this paper we propose a novel approach which combines the strengths of modern deep neural networks with the power of the watershed transform. Our model is simple, fast, accurate, and inherently handles an arbitrary number of instances per image with ease. 

%% file: review.tex
\section{A Review on the Watershed Transform}
 
We start our discussion with a review of the watershed transform, a well studied method in mathematical morphology.  
This technique is built on the fact that any greyscale image can be considered as a topographic surface. If we flood this surface from its minima while building barriers to prevent the merging of the waters coming from different sources, we effectively partition the image into different components or regions. These components are called catchment basins. The barriers or watershed lines then represent the boundaries between the different basins (i.e., boundaries between regions). 

\begin{figure}[t]
\vspace{-0.5cm}
\begin{center}
\begin{subfigure}[b]{0.4\textwidth} \centering \includegraphics[width=\textwidth]{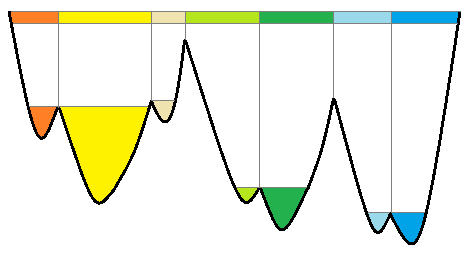} \end{subfigure} \\
(a) Traditional Watershed Energy
\begin{subfigure}[b]{0.4\textwidth} \centering \includegraphics[width=\textwidth]{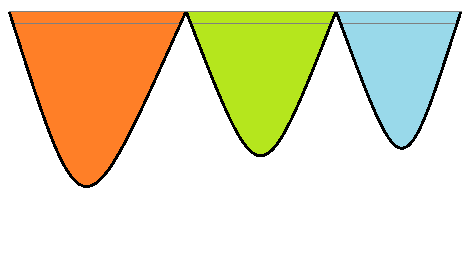} \end{subfigure} \\
\vspace{-1.cm}
(b) Our learned energy
\caption{Conceptual 1-D comparison between the traditional watershed transform and our deep watershed transform. The resulting instances are represented as colors across the top. }\label{fig:watershed_1d}
\end{center}
\end{figure}

This process is illustrated in the first row of Fig. \ref{fig:watershed_1d} for a one dimensional energy function. In this case, the watershed transform results in seven components, which are illustrated in different colors. 
Note that the traditional watershed transform tends to produce an over-segmentation of the image due to spurious small ridges which produce separate components. 
In this example, although there are 3 main components, the watershed transform over-segments the image because of small perturbations in the energy. 

Several algorithms have been developed to estimate the components. \cite{Beucher_1991_watershed} proposed an algorithm that iteratively fills the watershed landscape from each local minimum, adding dams wherever two neighbouring bodies of water meet. These dams define the segmentation boundaries. Additionally, \cite{Meyer_arxiv_watershed} details a number of alternative watershed transform algorithms, including topological distance, shortest path algorithms, spanning trees, and marker based methods. 

The watershed transform is typically applied to the image gradient, while the catchment basins correspond to homogeneous grey level regions in the image. 
However, estimating sharp gradients that represent the boundaries between different instances is a very challenging process. 
In the next section, we will show an alternative approach which directly learns to predict the energy landscape.

%% file: method.tex
\section{Deep Watershed Tranform}


In this section, we present our novel approach to instance level segmentation. In particular, we learn the energy of the watershed transform with a feed-forward neural network. The idea behind our approach is very intuitive. It consists of learning to predict an energy landscape such that 
each basin corresponds to a single instance, while all ridges are at the same height in the energy domain. As a consequence, the watershed cut corresponds to a single threshold in the energy, which does not lead to over segmentation. We refer the reader to the lower half of Fig. \ref{fig:watershed_1d} for an illustration of the desired energy.

\begin{figure*}[t]
\vspace{-0.5cm}
\begin{center}

\begin{subfigure}[b]{1.0\textwidth} \centering \includegraphics[width=\textwidth]{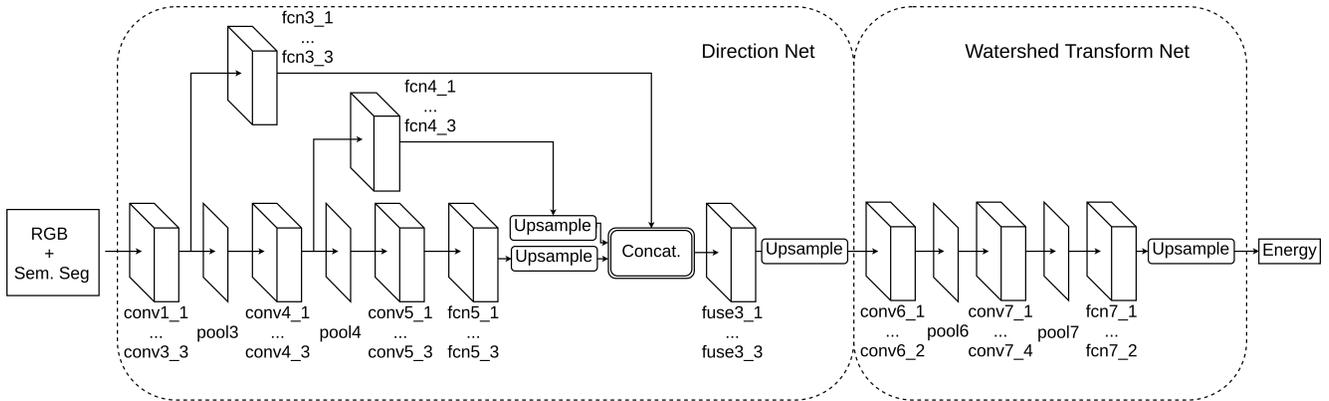} \end{subfigure}

\caption{Complete network architecture. The network takes the original RGB image gated by semantic segmentation and concatenated with the semantic segmentation as input, and produces the deep watershed transform energy map as the output. }\label{fig:network}
\end{center}
\end{figure*}

\begin{figure*}[t]
\vspace{-0.5cm}
\begin{center}

\begin{subfigure}[b]{0.24\textwidth} \centering \includegraphics[trim={0 0.0cm 0 0.cm},clip,width=\textwidth]{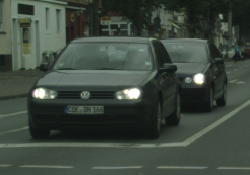}\caption{Input Image} \end{subfigure}
\begin{subfigure}[b]{0.24\textwidth} \centering \includegraphics[trim={0 0.0cm 0 0.cm},clip,width=\textwidth]{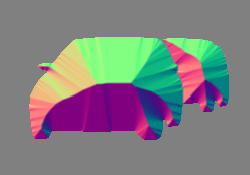}\caption{GT angle of $\vec{u}_p$}\end{subfigure} 
\begin{subfigure}[b]{0.24\textwidth} \centering \includegraphics[trim={0 0.0cm 0 0.cm},clip,width=\textwidth]{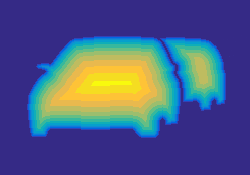}\caption{GT Watershed Energy}\end{subfigure} 
\begin{subfigure}[b]{0.24\textwidth} \centering \includegraphics[trim={0 0.0cm 0 0.cm},clip,width=\textwidth]{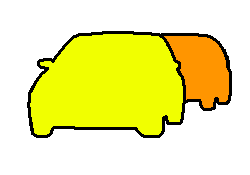}\caption{GT Instances} \end{subfigure} 

\begin{subfigure}[b]{0.24\textwidth} \centering \includegraphics[trim={0 0.0cm 0 0.cm},clip,width=\textwidth]{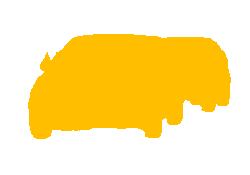} \caption{Sem. Segmentation of \cite{PSPNet}} \end{subfigure}
\begin{subfigure}[b]{0.24\textwidth} \centering \includegraphics[trim={0 0.0cm 0 0.cm},clip,width=\textwidth]{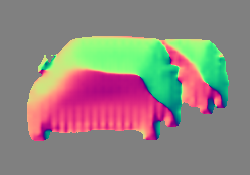}\caption{Pred. angle of $\vec{u}_p$} \end{subfigure}
\begin{subfigure}[b]{0.24\textwidth} \centering \includegraphics[trim={0 0.0cm 0 0.cm},clip,width=\textwidth]{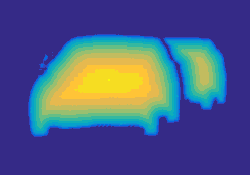}\caption{Pred. Watershed Transform} \end{subfigure}
\begin{subfigure}[b]{0.24\textwidth} \centering \includegraphics[trim={0 0.0cm 0 0.cm},clip,width=\textwidth]{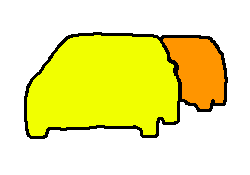} \caption{Pred. Instances} \end{subfigure}

\caption{Our network takes the RGB image (a) and the semantic segmentation (e) as input, and predicts a unit vector at each foreground pixel pointing directly away from the nearest boundary (f). Based on this, we then predict a modified watershed transform energy (g), upon which we perform cut at a fixed threshold to yield the final predictions (h).}\label{fig:watershed_intermediate}
\end{center}

\end{figure*}

Unfortunately, learning the energy landscape from scratch is a complex task. 
Therefore, we aid the network by defining an intermediate task, where we learn the direction of descent of the watershed energy. This is then passed through another set of network layers to learn the final energy. In principle, one can interpret this network as learning to perform the distance transform of each point within an object instance to the instance's boundary. Fig. \ref{fig:watershed_intermediate} shows an example of the input, intermediate results, and final output of our method. 
We refer the reader to Fig. \ref{fig:network} for an illustration of our network architecture. 

\subsection{Direction Network (DN)}
\label{subsec:direction_network}



Our approach leverages semantic segmentation as input to focus only on relevant areas of the image. Note that our network is agnostic to the choice of segmentation algorithm. In our paper, we use the semantic segmentation results from PSPNet \cite{PSPNet}.

The network takes as input the original RGB image gated by a binarized semantic segmentation, where all pixels that are not part of one of the semantic classes of interest are set to zero. The input image is augmented by adding the semantic segmentation as a fourth channel. Because the RGB values range from 0 to 255 for each channel before mean subtraction, we likewise scale the encoding of the semantic segmentation image such that the class labels are equally spaced numerically, and the variance is approximately equal to that of values in the RGB channels. 

To aid the model in producing an accurate energy landscape, we pre-train the overall network's first part (referenced here as the Direction Network, DN) to estimate the direction of descent of the energy at each pixel. 
We parameterize it with a unit vector pointing away from the nearest point on the object's instance boundary. 
This supervision gives a very strong training signal: the direction of the nearest boundary is directly encoded in the unit vector. 
Furthermore, a pair of pixels straddling an occlusion boundary between two objects will have target unit vectors pointing in opposite directions. Here, associating a pixel with the wrong object instance incurs the maximum possible angular error. This is true regardless of the shape of the objects, even when they are highly concave or elongated. This forces the network to learn very accurate boundary localization at the pixel level.

Let $D_\text{gt}(p)$ be the ground truth distance transform from a point to the boundary of the instance it belongs to. 
We define our ground truth targets as the normalized gradient $\vec{u}_p$ of this distance transform. More formally, 

\[
\vec{u}_{p, \text{gt}} =  
\frac{\nabla D_\text{gt}(p)}{|\nabla D_\text{gt}(p)|}
\]

Thus, the DN produces a two channel output at the input image's resolution representing the directional unit vector. It is important to note that the normalization layer at the output of the DN restricts the sum of each channel's squared output to be 1. This greatly reduces the difficulty of using the output of non-linear activation functions to model a scalar target. Additionally, phrasing the training target as a unit vector instead of an angle bypasses the problem of the equivalent angles 0$^{\circ}$ and 360$^{\circ}$ having a large numerical difference.

The feature extraction portion of the DN's architecture is inspired by VGG16 \cite{Simonyan_ICLR2015_VGG}. However, there are several important modifications. Direction prediction is a very precise task, as the output can vary greatly between neighboring pixels. Thus, it is critical to avoid losing spatial resolution. 
We utilize a modified version of the first 13 layers of VGG, where the third and fourth pooling layers are changed to average pooling while the fifth pooling layer is removed. 
To preserve spatial resolution, we exploit a high-capacity, multi-scale information aggregation scheme inspired by popular methods in semantic segmentation  \cite{Ghiasi_ECCV2016_LRR,Yu_ICLR2016_dilation,Long_CVPR2015_FCN}. In particular, the outputs of {\it conv3}, {\it conv4}, and {\it conv5} individually undergo a $5 \times 5$ convolution followed by two $1 \times 1$ convolutions. After this, the outputs of the second and third paths are upsampled to the resolution of the first path. The resulting feature volumes from the three paths are concatenated. This undergoes an additional set of three $1 \times 1$ convolutions, before being upsampled to the input resolution. 

\subsection{Watershed Transform Network (WTN)}\label{subsec:watershed_network}

The second half of the overall network takes as input the 2-channel unit vector map, and produces a discretized modified watershed transform map with $K=16$ possible energy values. In this case, bin 0 corresponds to background or regions within 2 pixels of an instance boundary and is referred to as having an energy level of 0. Meanwhile, higher numbered bins with higher energy levels correspond to regions in the interior of object instances. The bins are chosen to maximize the binning resolution of energy levels near zero (to facilitate accurate cutting), while achieving an  approximate balance between the total numbers of pixels within each class. 

The WTN is a fairly generic CNN, with an emphasis on reasoning with high spatial accuracy and resolution. The architecture is shown in the right half of Fig. \ref{fig:network}. In particular, the network consists of two $5 \times 5$ convolutional filter blocks each followed by a $2 \times 2$ average pooling, before undergoing a set of $1 \times 1$ convolutions and upsampling to the input image resolution. 


\subsection{Network Training}


We first pre-train our DN and WTN networks. This process is followed by end-to-end training of the whole network. 

\vspace{-3.5mm}

\paragraph{Direction Network pre-training: } We pre-train the network using the mean squared error in the angular domain as loss:
\[
l_\text{direction} = \sum_{p \in \mathcal{P}_\text{obj}} w_p \|\cos^{-1}<\vec{u}_{p, \text{GT}}, \vec{u}_{p, \text{pred}}>\|^2 
\]
where $\mathcal{P}_\text{obj}$ is the set of all pixels belonging to a semantic class with object instances, and $w_p$ is a weighting factor proportional to the inverse of the square root of the object instance's area. We use the mean squared angular error as opposed to the commonly used cosine distance, as large angular errors (occuring when the network associates a pixel with the wrong object instance) should incur significant larger penalties than small angular errors (which have little impact).

This network is pre-trained using the original input images gated by the ground truth semantic segmentation and the ground truth unit vectors $\vec{u}_\text{GT}$. The lower layers (conv1 to conv5) are initialized with VGG16 \cite{Simonyan_ICLR2015_VGG}. The weights of the upper layers are initialized randomly according to the Xavier initialization scheme \cite{Xavier}. However, the intialization variance of the weights of the final layer of each prediction branch is set manually such that the variance of each branch's output is of the same order of magnitude before concatenation. This encourages the network to consider the output of each branch. We use the ADAM optimizer to train the network for 20 epochs with a batch size of 4, with a constant learning rate of 1e-5 and L2 weight penalty of 1e-5.

\vspace{-3.5mm}

\paragraph{Watershed Network pre-training: } The network is trained using a modified cross-entropy loss 
\[
\begin{split}
l_\text{watershed} = \sum_{p \in \mathcal{P}_\text{obj}}  \sum_{k=1}^{K} w_p c_k(\bar{t}_{p,k} \log \bar{y}_{p,k} +  t_{p,k}\log y_{p,k})
\end{split}
\]
where $t_{p,k}$ is the $k$-th element of pixel $p$'s one-hot target vector, $y_{p,k}$ is the $k$-th channel of the network output at $p$, $w_p$ is a coefficient to adjust the importance of smaller objects (as defined in the DN pre-training section), and $c_k$ is a scaling constant specific to each discretization class. Because the single energy level cut occurs at just above zero energy (i.e., near object instance boundaries), an accurate estimation of pixels at low energy levels is of crucial importance. Thus, the set $\{c_k\}$ is selected to be in increasing order. In this case, any errors 
of low energy levels are assigned a greater level of penalty, which encourages the network to focus on boundary regions. 
We pre-train the network using the ground truth semantic segmentation and ground truth direction predictions as input, and the discretized ground truth modified watershed transform as target. All weights are initialized using Xavier initialization \cite{Xavier}. We train the network for 25 epochs using the ADAM optimizer. A batch size of 6, constant learning rate of 5e-4, and a L2 weight penalty of 1e-6 are used. 

\vspace{-3.5mm}

\paragraph{End-to-end fine-tuning: } We cascaded the pre-trained models for the DN and WTN and fine-tuned the complete model for 20 epochs using the RGB image and semantic segmentation output of PSPNet as input, and the ground truth distance transforms as the training target. We use a batch size of 3, constant learning rate of 5e-6, and a L2 weight penalty of 1e-6.

\subsection{Energy Cut and Instance Extraction}\label{subsec:inference}

We cut our watershed transform output at energy level 1 (0 to 2 pixels from boundary) for classes with many smaller objects (\textit{person}, \textit{rider}, \textit{bicycle}, \textit{motorcycle}), and level 2 (3 to 4 pixels from boundary) for classes with larger objects ({\textit{car}, \textit{bus}, \textit{truck}, \textit{train}}). Following this, the instances are dilated using a circular structuring element whose radius equals the maximum erosion at the energy level of the cut. This offsets the boundary erosion from the non-zero threshold. The connected components are identified in the resulting image, directly yielding the proposed instances. The proposals are further refined by basic hole-filling. Finally, we remove small, spurious instances. 


%% file: results.tex

\section{Experimental Evaluation}

In this section, we evaluate  our approach on the challenging Cityscapes Instance-Level Semantic Labelling Task \cite{Cityscapes}. The official benchmark test and validation set results are found in Tables \ref{maintestset} and \ref{classtestset}. We then perform ablation studies with the validation set to examine the performance of various aspects of our model.

\vspace{-3.7mm}

\begin{table*}[t]
\centering
\begin{tabular}{|l||l|l|l|l||l|l|}
\hline
\multicolumn{1}{|c||}{\multirow{2}{*}{Method}} & \multicolumn{4}{c||}{Cityscapes Test Set}                                                                                     & \multicolumn{2}{c|}{Cityscapes Val Set}              \\ \cline{2-7} 
\multicolumn{1}{|c||}{}                        & \multicolumn{1}{c|}{\textbf{AP}} & \multicolumn{1}{c|}{AP 50\%} & \multicolumn{1}{c|}{AP 100m} & \multicolumn{1}{c||}{AP 50m} & \multicolumn{1}{c|}{AP} & \multicolumn{1}{c|}{muCov} \\ \hline \hline 
van den Brand et al. \cite{Brand_ACCV2016_DeepContours}      & 2.3\%                              & 3.7\%                        & 3.9\%                        & 4.9\%                       & -                       & -                          \\ \hline
Cordts et al. \cite{Cityscapes}     & 4.6\%                            & 12.9\%                       & 7.7\%                        & 10.3\%                      & -                       & -                          \\ \hline
Uhrig et al. \cite{Uhrig_GCPR2016_Pixel}    & 8.9\%                            & 21.1\%                       & 15.3\%                       & 16.7\%                      & 9.9\%                   & -                          \\ \hline
\textbf{Ours}            & \textbf{19.4\%}                  & \textbf{35.3\%}              & \textbf{31.4\%}              & \textbf{36.8\%}             & \textbf{21.2\%}         & \textbf{68.0}\%                     \\ \hline
\end{tabular}
\caption{Cityscapes instance segmentation results using metrics defined in \cite{Cityscapes} for AP and \cite{mucov} for muCov.}\label{maintestset}
\end{table*}
\mbox{}

\begin{table*}[]
\centering
\begin{tabular}{|l||l|l|l|l|l|l|l|l|}
\hline
Method        & Person          & Rider           & Car             & Truck           & Bus             & Train           & Motorcycle     & Bicycle        \\ \hline \hline 
van den Brand et al. \cite{Brand_ACCV2016_DeepContours}         & -             & -             & 18.2\%          & -             & -             & -             & -            & -            \\ \hline
Cordts et al. \cite{Cityscapes}         & 1.3\%           & 0.6\%           & 10.5\%          & 6.1\%           & 9.7\%           & 5.9\%           & 1.7\%          & 0.5\%          \\ \hline
Uhrig et al. \cite{Uhrig_GCPR2016_Pixel}         & 12.5\%          & 11.7\% & 22.5\%          & 3.3\%           & 5.9\%           & 3.2\%           & 6.9\%          & 5.1\% \\ \hline
\textbf{Ours} & \textbf{15.5\%} & \textbf{14.1\%} & \textbf{31.5\%} & \textbf{22.5\%} & \textbf{27.0\%} & \textbf{22.9\%} & \textbf{13.9\%} & \textbf{8.0\%}          \\ \hline
\end{tabular}
\caption{Cityscapes instance segmentation class specific test set AP scores using metrics defined in \cite{Cityscapes}.}\label{classtestset}
\end{table*}
\mbox{}

\vspace{-3.7mm}

\paragraph{Dataset:}
The Cityscapes Instance Labelling Task contains 5000 finely annotated street scene images taken by a vehicle-mounted camera. Each image has a resolution of 2048x1024 pixels. 
Unlike other commonly used datasets for instance segmentation (e.g., Pascal VOC \cite{Pascal2012}, BSDS500 \cite{BSDS500}, and CVPPP \cite{CVPPP}) Cityscapes has a large number of scenes involving dozens of instances  with large degrees of occlusions at vastly different scales. 
Eight of the semantically labelled categories have instance-level labelling. We refer the reader to Table \ref{cityscapesdataset} for a summary of the statistics of the object instances in this dataset. 
Note that while the \textit{car} and \textit{people} classes have significant numbers of instances, the other six classes are rather uncommon. As a result, the rare classes have far less training data, and are thus much more challenging. 
We use the official training, validation, and testing set splits, with 2975, 500, and 1525 images, respectively. 

\vspace{-3.7mm}

\paragraph{Metric:} We report the metrics employed by the Cityscapes leaderboard. Several variants of the instance-level average precision score are used. This is calculated by finding the precision at various levels of intersection-over-union (IoU) between predicted instances and ground truth instances, ranging from 50\% to 95\%. The main score is the Average Precision (AP). Additionally, there are three minor scores: AP at 50\% overlap, AP of objects closer than 50m, and AP of objects closer than 100m. Methods are ranked by the AP score over all classes (mean AP). 
Note that AP is a detection score, which does not penalize overlapping instances or a large number of predictions, as long as they are ranked in the proper order. 
Thus, it places approaches (like ours) which predict a single instance label (or background) per pixel at a disadvantage, while favoring detection-based methods. Despite this, we use these metrics to be consistent with the the evaluation of the state-of-the-art. For the \{\textit{bus},\textit{truck},\textit{train}\} classes, we order the instances by simply averaging the semantic segmentation's output confidence within each instance to somewhat counteract the errors by semantic segmentation. For all other classes, we use a random ordering. Note that sophisticated ranking techniques can be used to further improve the score. However, the segmentation quality remains this same. This again highlights a shortcoming of using AP as the metric.  

In addition to AP, we report the mean weighted coverage score introduced in \cite{mucov} on the validation set. This metric enforces a single instance label per pixel, and is therefore more suitable for evaluating our approach. We hope that future work will likewise report this score when applicable. 

\vspace{-4.7mm}

\begin{table}[]
\centering
\begin{tabular}{|p{1.45cm}||p{1.0cm}|p{1.45cm}|p{1.25cm}|p{1.25cm}|}
\hline
Category   & Total Pixels & Total Instances & Average Size & Instances / Image \\ \hline \hline  
Person     & 6.6e+7      & 17900           & 3691         & 6.0               \\ \hline
Rider      & 7.4e+6      & 1754            & 4244         & 0.6               \\ \hline
Car        & 3.8e+8      & 26929           & 14233        & 9.1               \\ \hline
Truck      & 1.5e+7      & 482             & 30648        & 0.2               \\ \hline
Bus        & 1.3e+7      & 379             & 34275        & 0.1               \\ \hline
Train      & 3.2e+6      & 168             & 19143        & 0.1               \\ \hline
Motorcycle & 5.4e+6      & 735             & 7298         & 0.2               \\ \hline
Bicycle    & 1.6e+7      & 3655            & 4332         & 1.2               \\ \hline
\end{tabular}
\caption{Statistics of the training set of the cityscapes instance level segmentation. Average size is in pixels.}\label{cityscapesdataset}
\end{table}
\mbox{}

\begin{table}[]
\centering
\begin{tabular}{|l||l|l|}
\hline
Sem. Seg. Src.         & Sem. Seg. IoU    & Inst. Seg. AP  \\ \hline \hline 
LRR \cite{Ghiasi_ECCV2016_LRR}  & 71.8\%  & 20.2\% \\ \hline
PSPNet \cite{PSPNet}       & \textbf{80.2}\% & \textbf{21.2\%} \\ \hline
\end{tabular}
\caption{Comparison of instance segmentation performance on the validation set with various semantic segmentation sources. Semantic segmentation IoU Class scores \cite{Cityscapes} are also provided.}\label{psp_lrr}
\end{table}

\begin{table*}[]
\centering
\setlength\tabcolsep{4.5pt}
\begin{tabular}{|p{3.5cm}||l|l|l|l|l|l|l|l|l|}
\hline
Method                 & Average & Person  & Rider   & Car     & Truck   & Bus     & Train   & Motorcycle & Bicycle \\ \hline \hline  
Ours    & 20.50\% & 15.50\%  & 13.80\%  & 33.10\% & 23.60\%  & 39.90\%  & 17.40\%  & 11.90\%      & 8.80\%   \\ \hline
Ours + Ordering    & 21.24\% & 15.50\%  & 13.80\%  & 33.10\% & 27.10\%  & 45.20\%  & 14.50\%  & 11.90\%      & 8.80\%  \\ \hline
Ours + Oracle Ordering & \textbf{27.58\%} & \textbf{20.60\%} & \textbf{18.70\%} & \textbf{40.10\%} & \textbf{31.50\%} & \textbf{50.60\%} & \textbf{28.30\%} & \textbf{17.40\%}    & \textbf{13.40\%}  \\ \hline
\end{tabular}
\caption{Comparison of AP scores with various instance ordering techniques using the validation set.}\label{ordering}
\end{table*}
\mbox{}

\vspace{-3.7mm}

\paragraph{Comparison to the state-of-the-art:} We show the instance segmentation test set scores in Table \ref{maintestset}. Additionally, we show the class-specific AP scores in Table \ref{classtestset}. It is evident that we achieve a large improvement over the  state-of-the-art in all semantic classes. Moreover, we do not use depth information to train our model, unlike \cite{Uhrig_GCPR2016_Pixel}. 

\vspace{-3.7mm}

\paragraph{Analysis of the intermediate training target: } Our final network is the result of the end-to-end fine-tuning of two pre-trained sub-networks (DN and WTN). Fig. \ref{fig:watershed_intermediate} (f) shows the output of the DN after finetuning. It is evident that the fine-tuned model retained the direction prediction as an intermediate task. This suggests that the intermediate training target is effective. 

\vspace{-3.7mm}

\paragraph{Influence of semantic segmentation: } While we elected to use PSPNet \cite{PSPNet} as our semantic segmentation source, we additionally demonstrate that our method is able to use other sources. Table \ref{psp_lrr} shows the use of LRR \cite{Ghiasi_ECCV2016_LRR} for semantic segmentation. Using the same pre-trained DN and DTN models, we perform end-to-end fine-tuning using LRR as semantic segmentation. We note that the performance of our model improves with better semantic segmentation. Thus, future advances in segmentic segmentation methods can further improve the performance of our approach. 

\vspace{-3.7mm}

\paragraph{Confidence score estimate: } As mentioned, the AP score calculated by the Cityscapes benchmark requires a confidence score for each instance. For the \{\textit{bus}, \textit{truck}, \textit{train}\} set, we produce a weak ranking based on semantic segmentation softmax confidence. Instances of all other classes are randomly ranked. Table \ref{ordering} explores the impact of various ordering schemes. We compare our ordering with random for all semantic classes, as well as optimal ordering using oracle IoU. We see that ordering using oracle IoU can increase our model's performance by 6.34\%. Note, however, that this has no impact on the actual quality of proposed object instances, which remain the same. 
This shows the necessity of a different metric such as muCov \cite{mucov} that can evaluate segmentation-based approaches fairly. 

\input{qualitative}




\vspace{-3.7mm}

\paragraph{Qualitative Results:}
Fig. \ref{fig:qualitative} depicts visualizations of sample results on the validation set, which is not used as part of training. It is evident that our model produces very high quality instance segmentation results. In these results, predicted and ground truth object instances only share the same color if they have greater than 50\% IoU. 

\vspace{-3.7mm}

\paragraph{Failure Modes:}
Our model has several weaknesses. Some images in Fig. \ref{fig:qualitative} demonstrate these cases. The first issue is that the current formulation of our method does not handle objects that have been separated into multiple pieces by occlusion. This is most obvious in the 3rd image from the bottom in Fig. \ref{fig:qualitative} as the far-right vehicle is bisected by the street sign pole, and the bicycle in the right part of the image just above. The resulting pieces are not merged into one component. This is a drawback of most bottom-up grouping approaches. The second issue are cases where two objects sharing an occlusion boundary are mostly but not fully separated by a line of low energy. This is seen in the right-most vehicle in the 11th image. We anticipate that a combination of our method with top-down reasoning approaches will greatly alleviate these two issues. 

Because we rely upon correct semantic segmentation, errors from this  (such as identifying a train as a bus) cannot be fixed by our method. This is clearly shown by the truck in the last example. A possible solution could be to use semantic segmentation as soft gating, or to reason about semantic and instance segmentation jointly. 
Finally, some very complex scenes such as some subgroups of people on the left in the second to fourth example are incorrectly separated by our model, and are fused together. 

%% file: qualitative.tex
\begin{figure*}[]
\begin{center}
\begin{subfigure}[b]{0.22\textwidth} \centering \includegraphics[trim={0 3cm 0 1.25cm},clip,width=\textwidth]{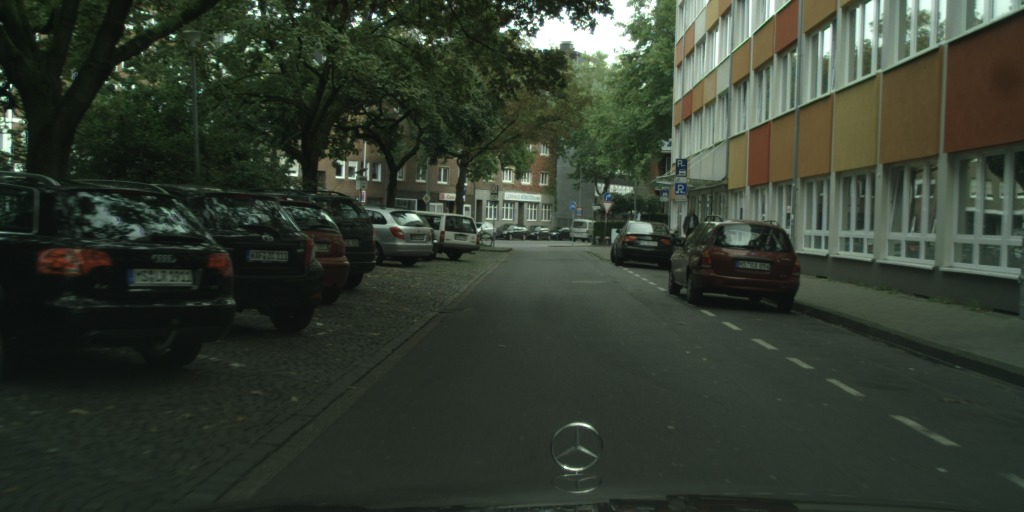}\end{subfigure}
\begin{subfigure}[b]{0.22\textwidth} \centering \includegraphics[trim={0 3cm 0 1.25cm},clip,width=\textwidth]{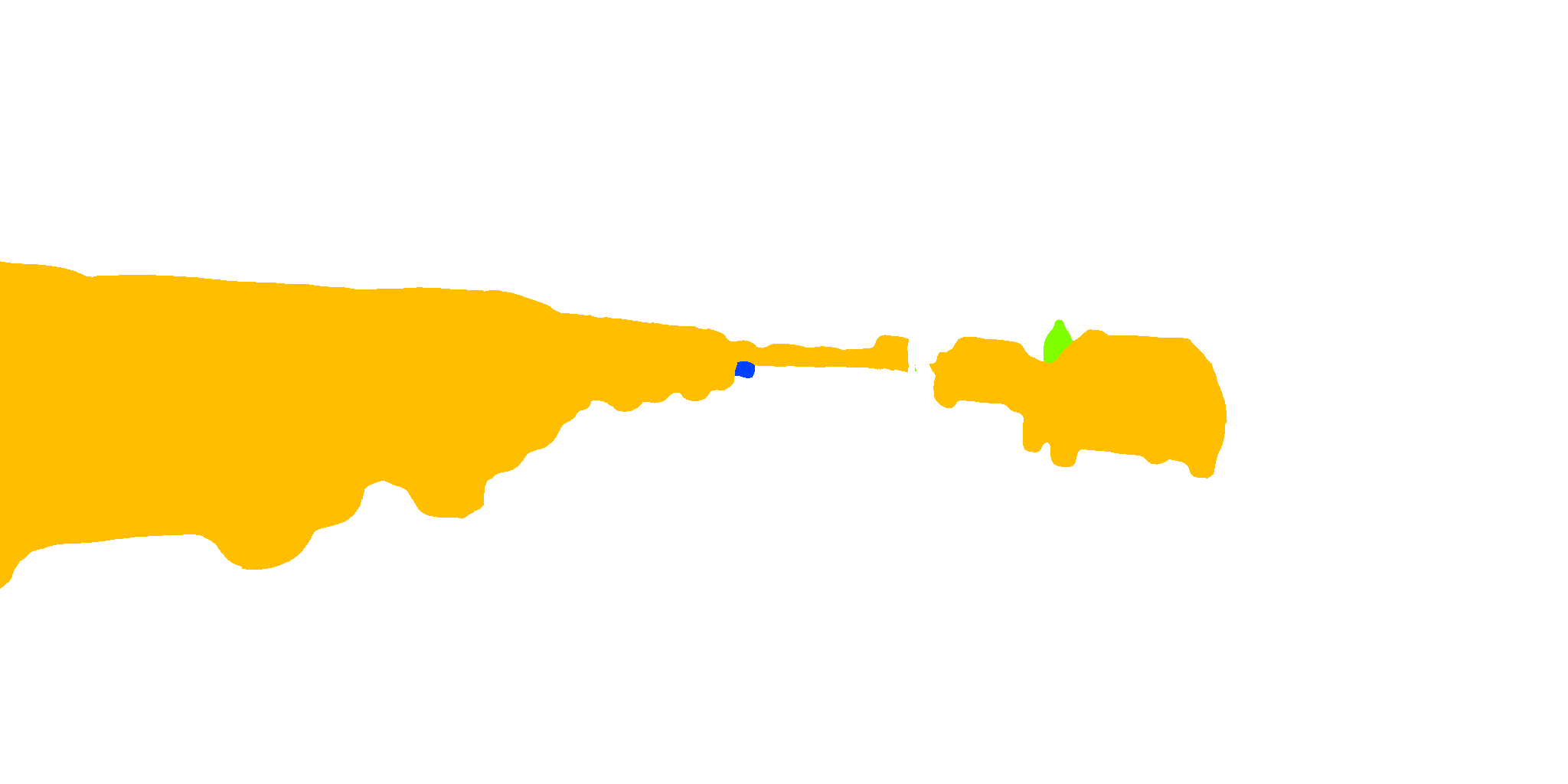}\end{subfigure}
\begin{subfigure}[b]{0.22\textwidth} \centering \includegraphics[trim={0 3cm 0 1.25cm},clip,width=\textwidth]{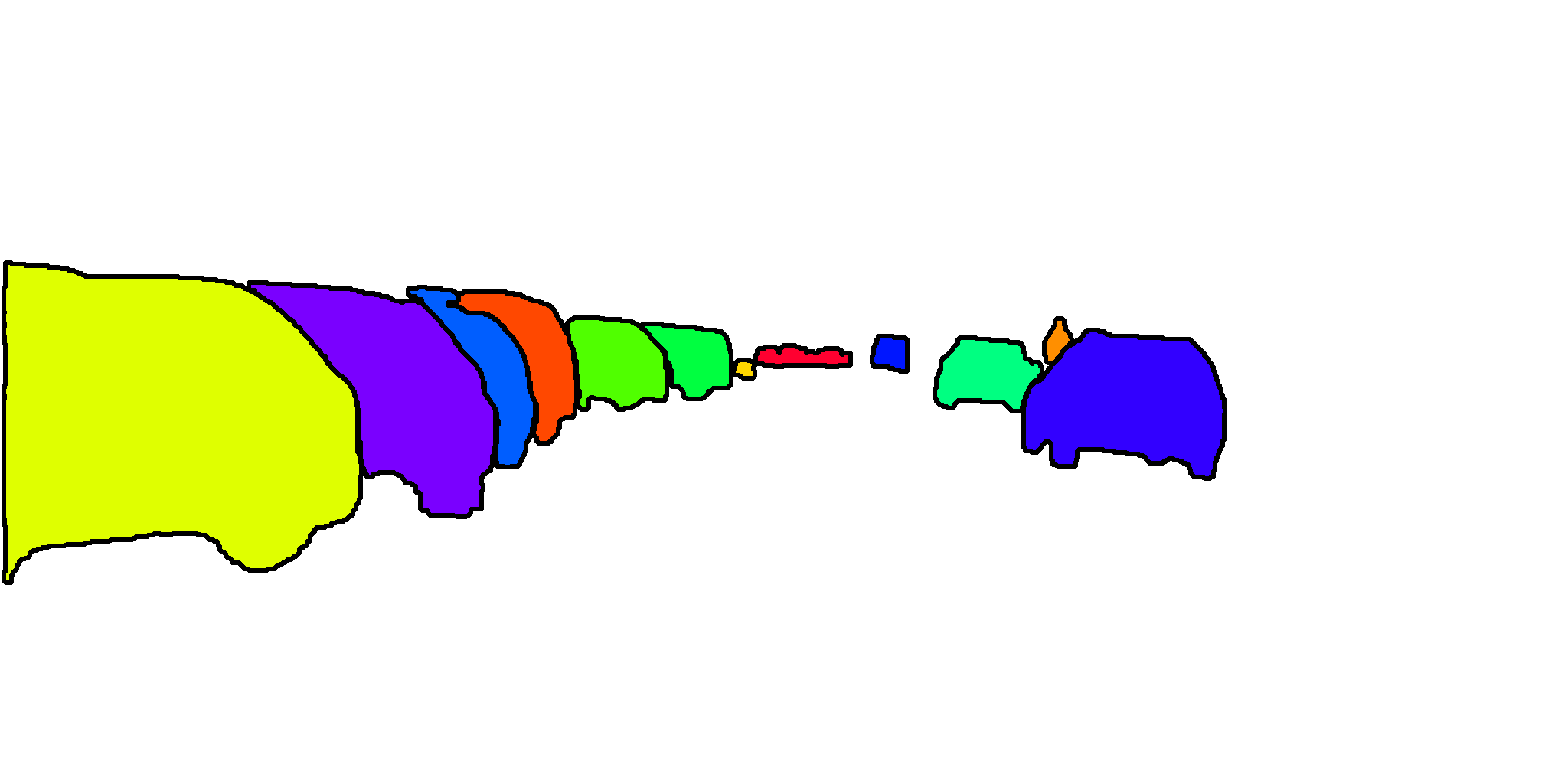}\end{subfigure}
\begin{subfigure}[b]{0.22\textwidth} \centering \includegraphics[trim={0 3cm 0 1.25cm},clip,width=\textwidth]{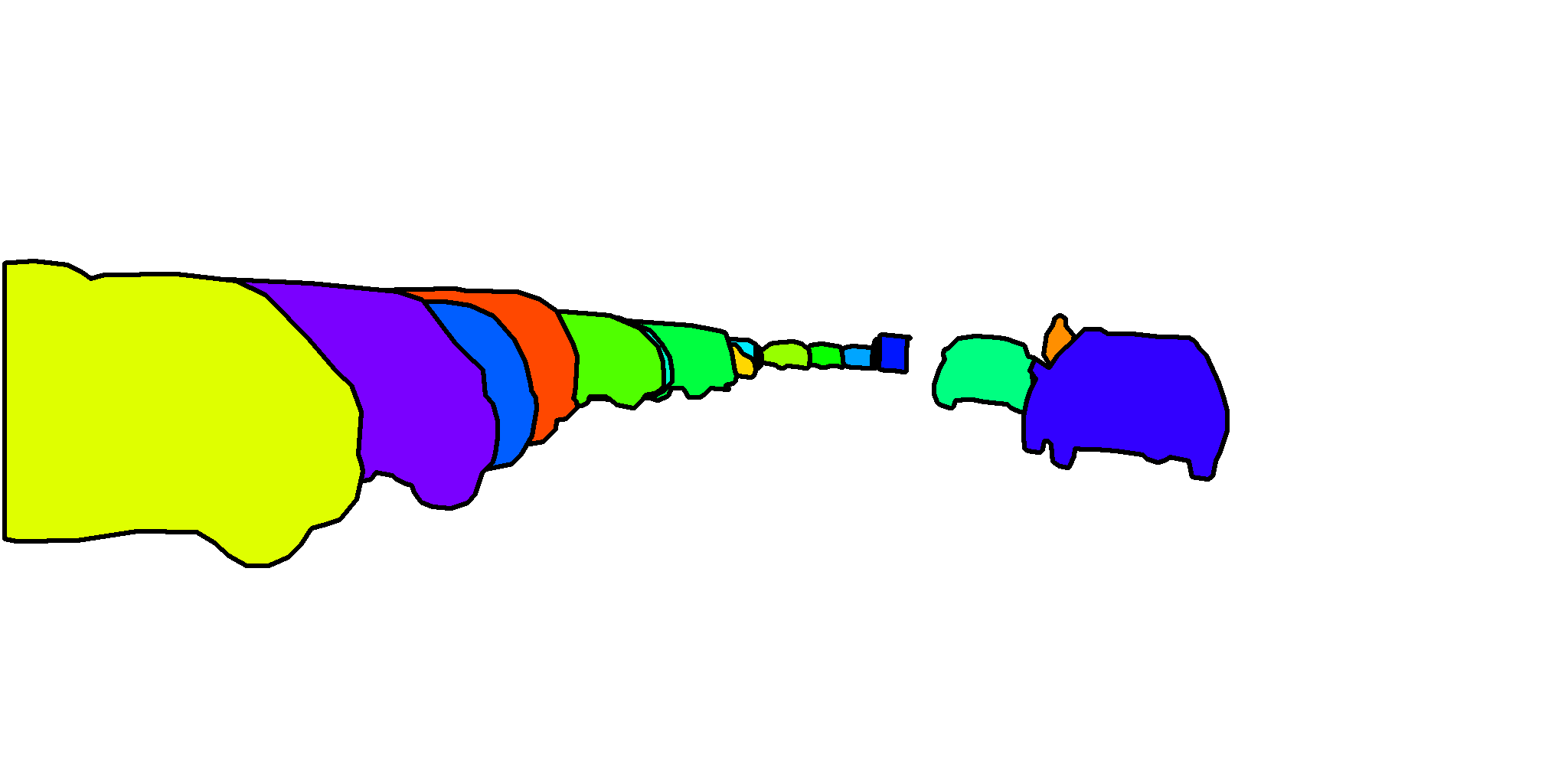}\end{subfigure}

\vspace{0.10cm}

\begin{subfigure}[b]{0.22\textwidth} \centering \includegraphics[trim={0 3cm 0 1.25cm},clip,width=\textwidth]{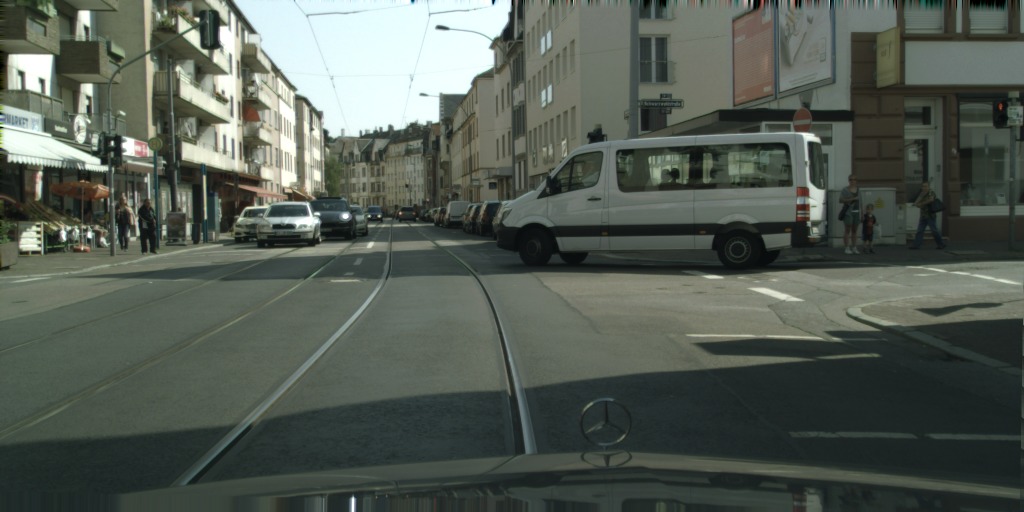}\end{subfigure}
\begin{subfigure}[b]{0.22\textwidth} \centering \includegraphics[trim={0 3cm 0 1.25cm},clip,width=\textwidth]{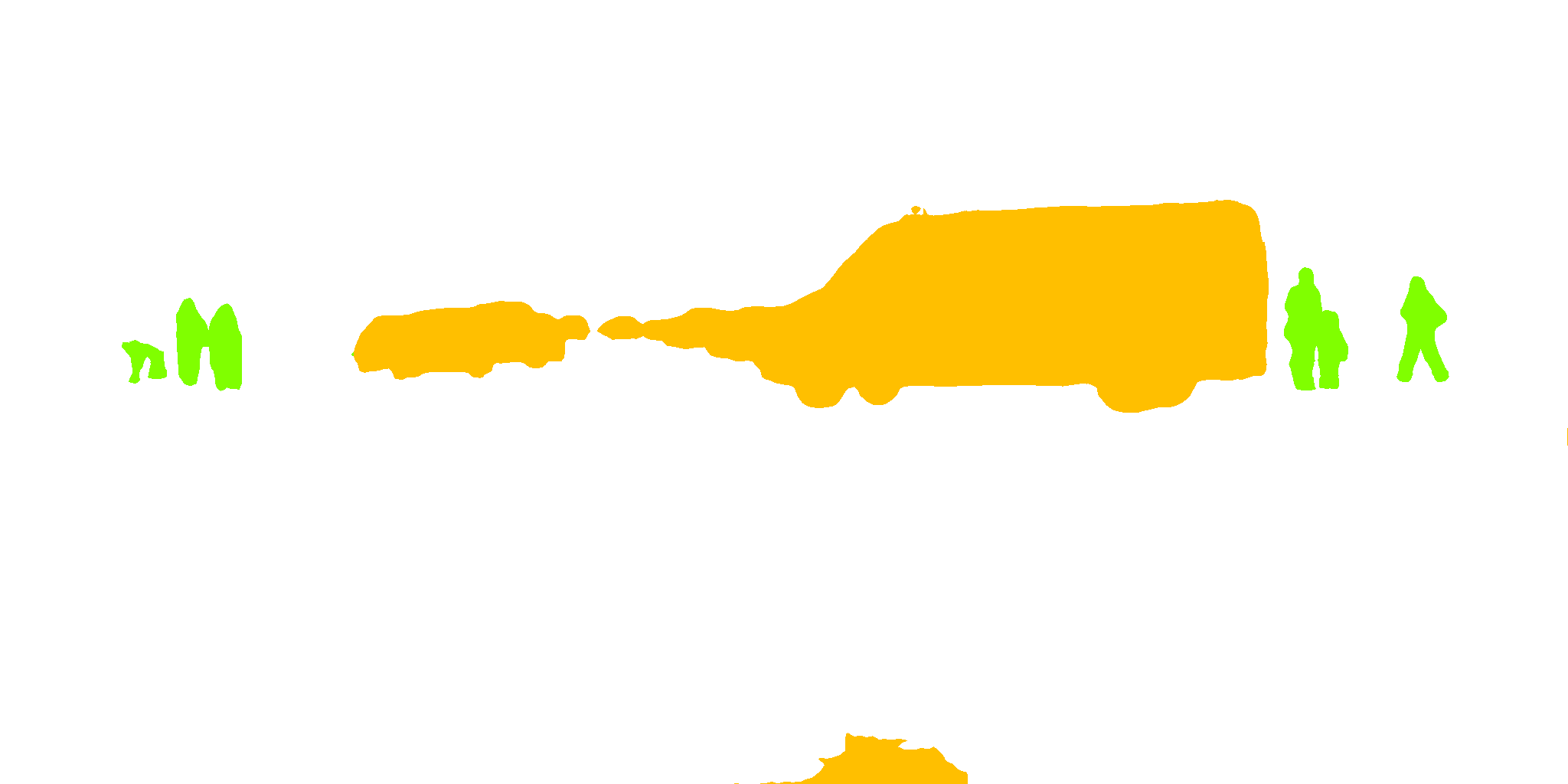}\end{subfigure}
\begin{subfigure}[b]{0.22\textwidth} \centering \includegraphics[trim={0 3cm 0 1.25cm},clip,width=\textwidth]{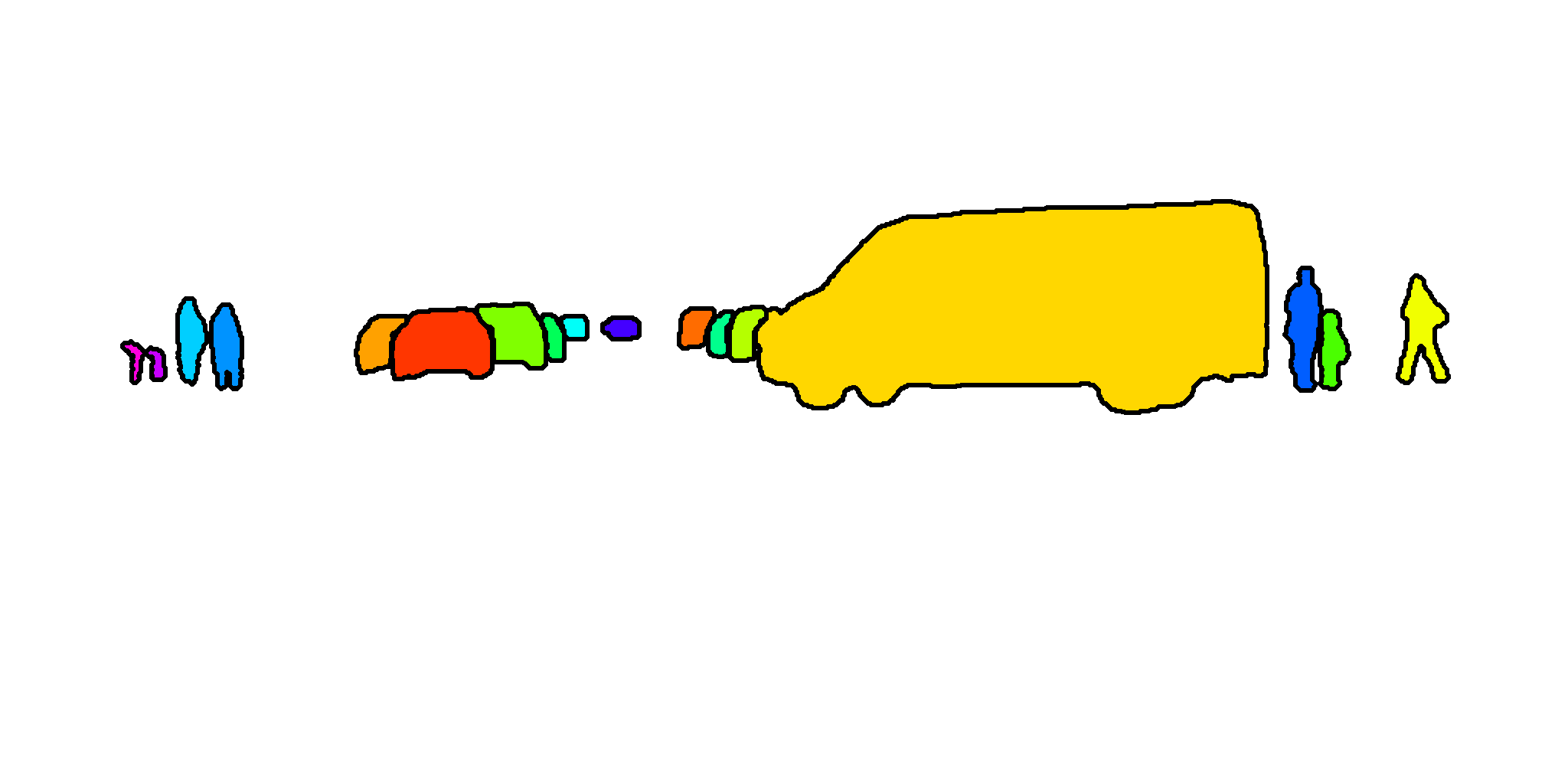}\end{subfigure}
\begin{subfigure}[b]{0.22\textwidth} \centering \includegraphics[trim={0 3cm 0 1.25cm},clip,width=\textwidth]{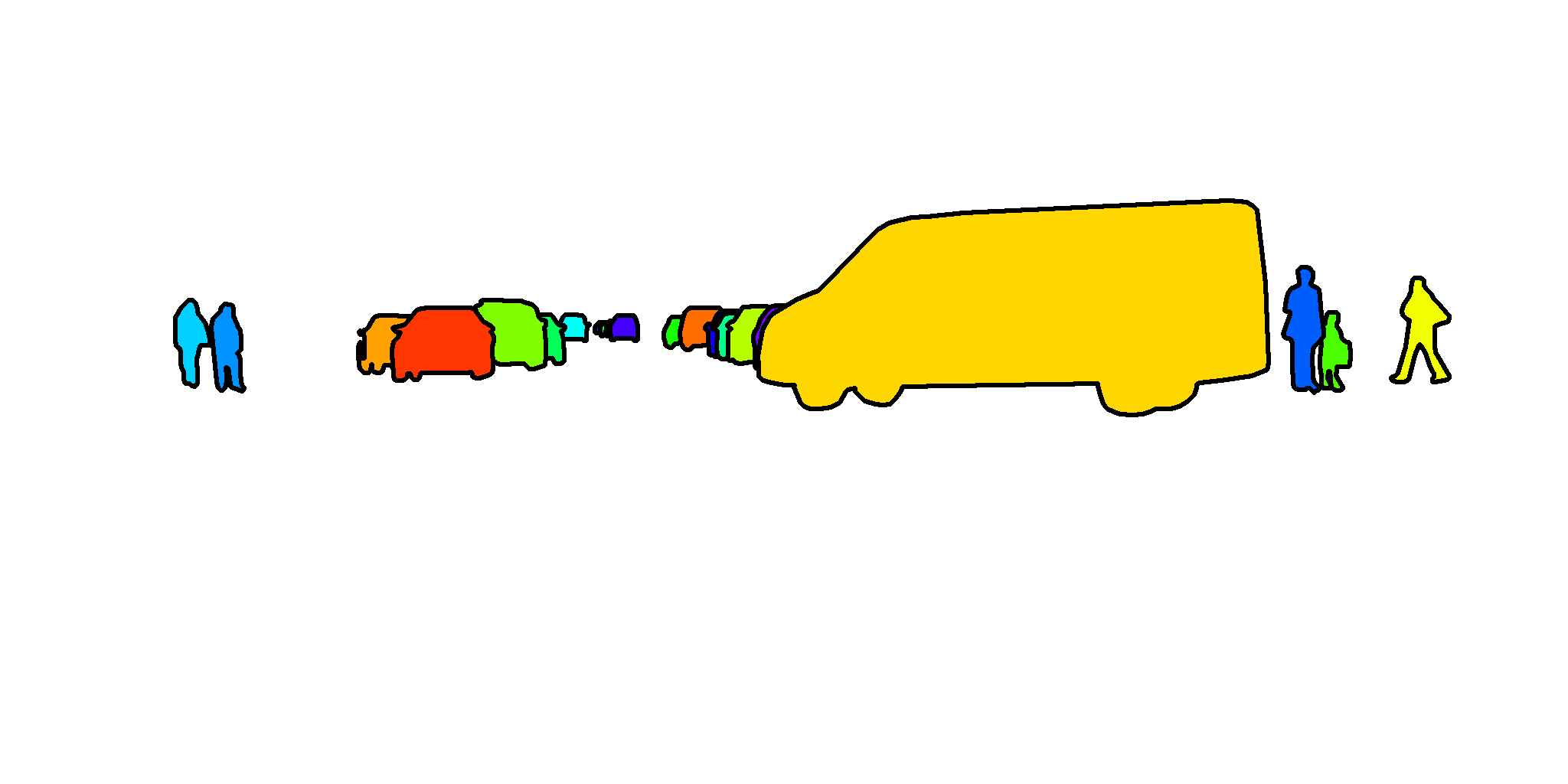}\end{subfigure}

\vspace{0.10cm}

\begin{subfigure}[b]{0.22\textwidth} \centering \includegraphics[trim={0 3cm 0 1.25cm},clip,width=\textwidth]{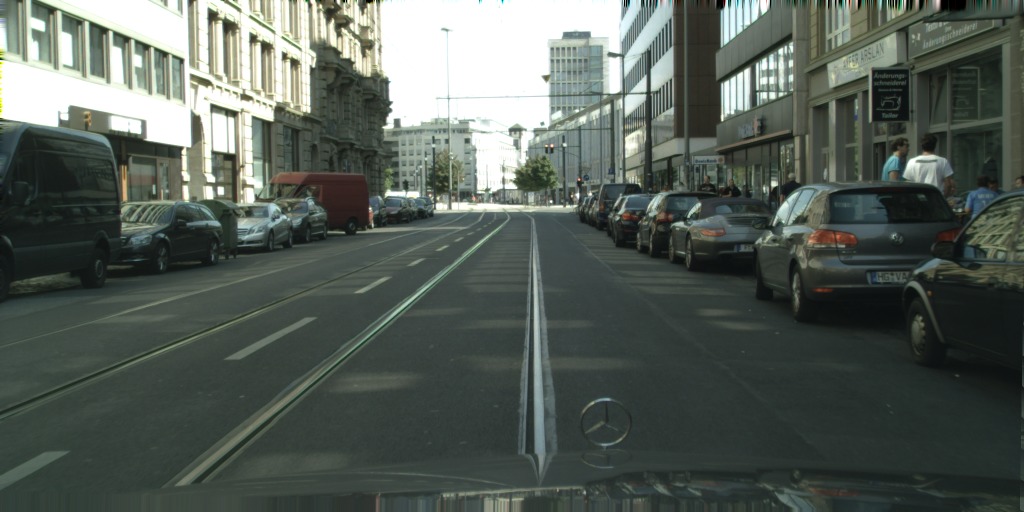}\end{subfigure}
\begin{subfigure}[b]{0.22\textwidth} \centering \includegraphics[trim={0 3cm 0 1.25cm},clip,width=\textwidth]{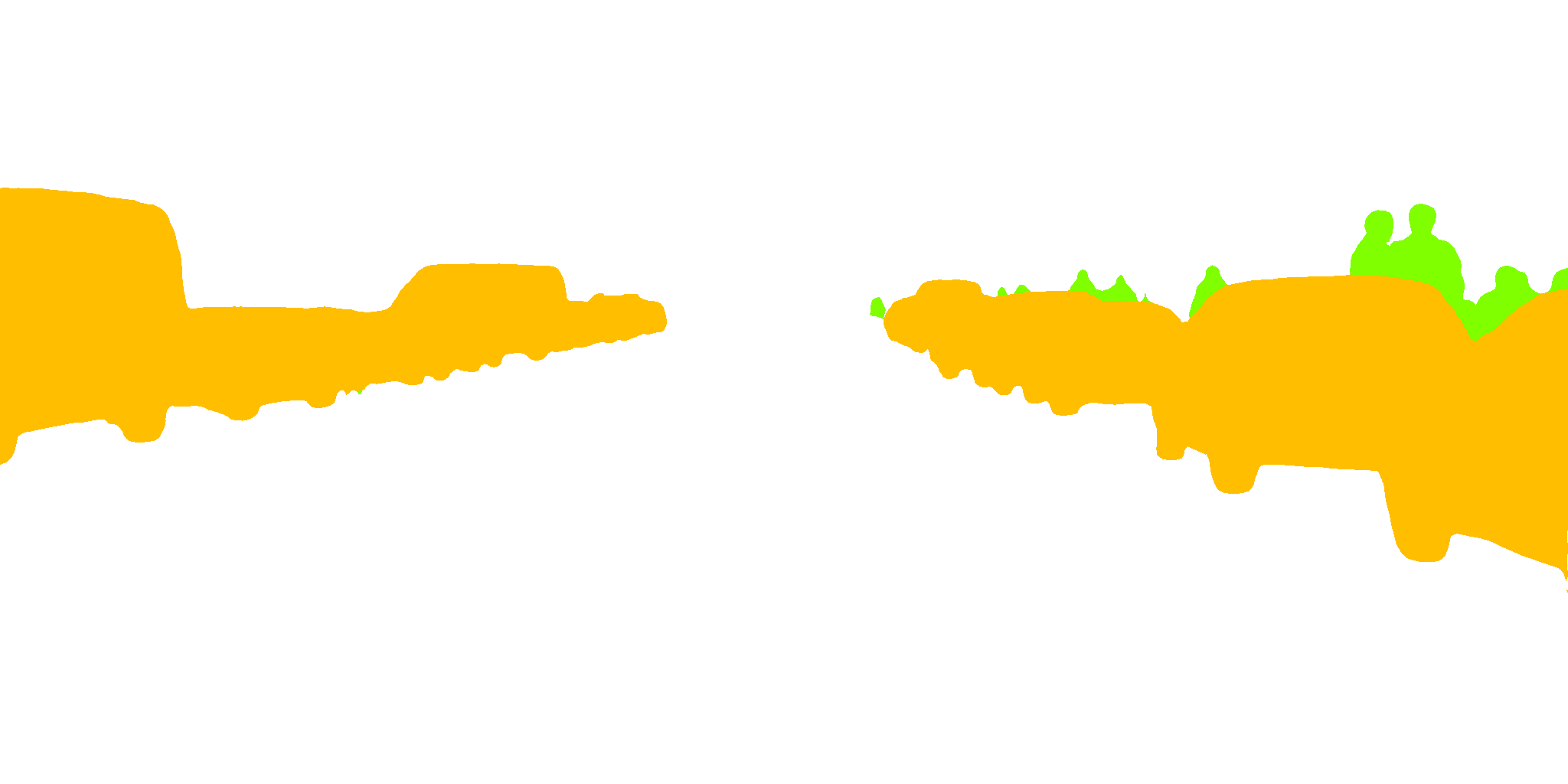}\end{subfigure}
\begin{subfigure}[b]{0.22\textwidth} \centering \includegraphics[trim={0 3cm 0 1.25cm},clip,width=\textwidth]{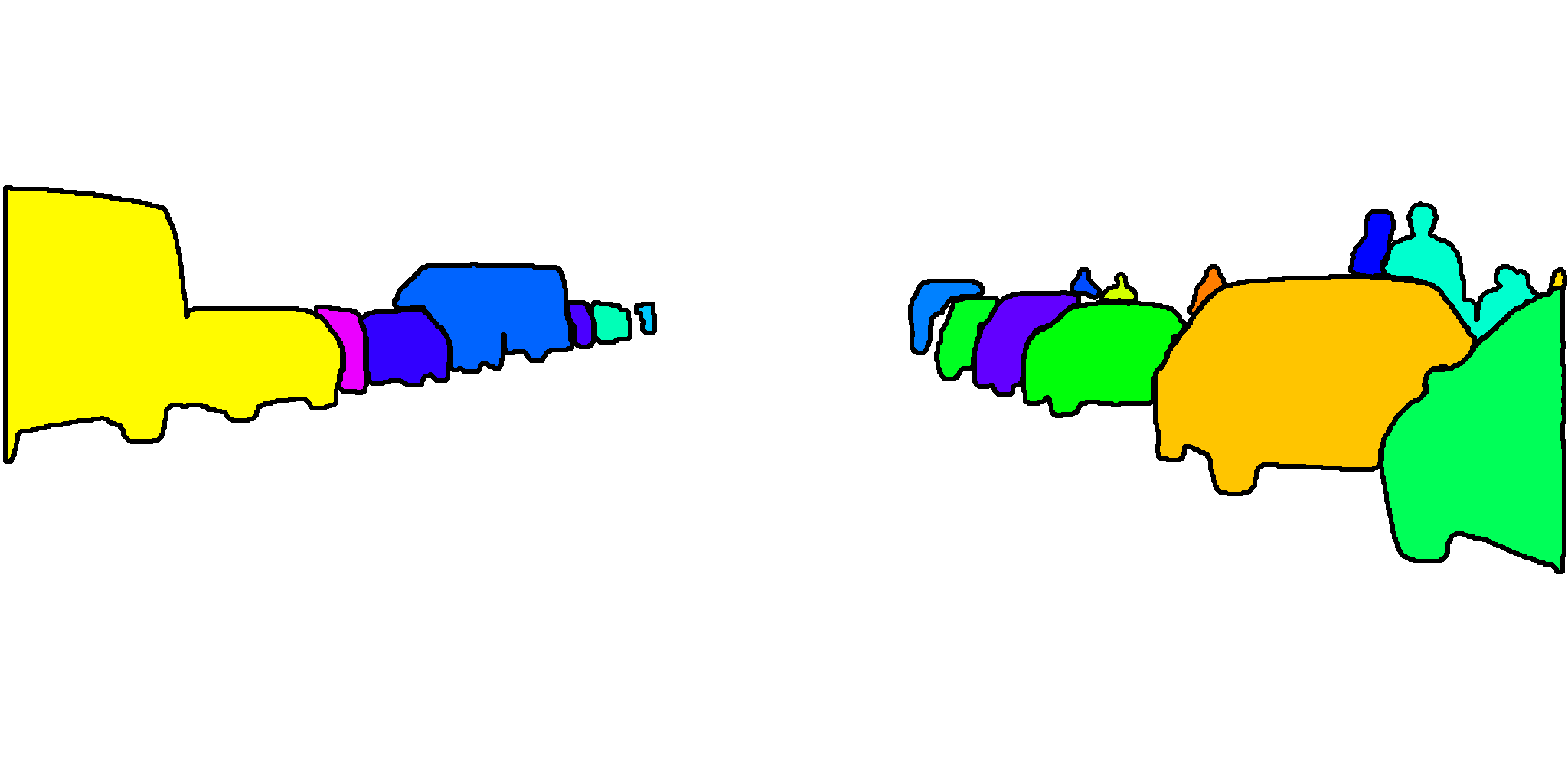}\end{subfigure}
\begin{subfigure}[b]{0.22\textwidth} \centering \includegraphics[trim={0 3cm 0 1.25cm},clip,width=\textwidth]{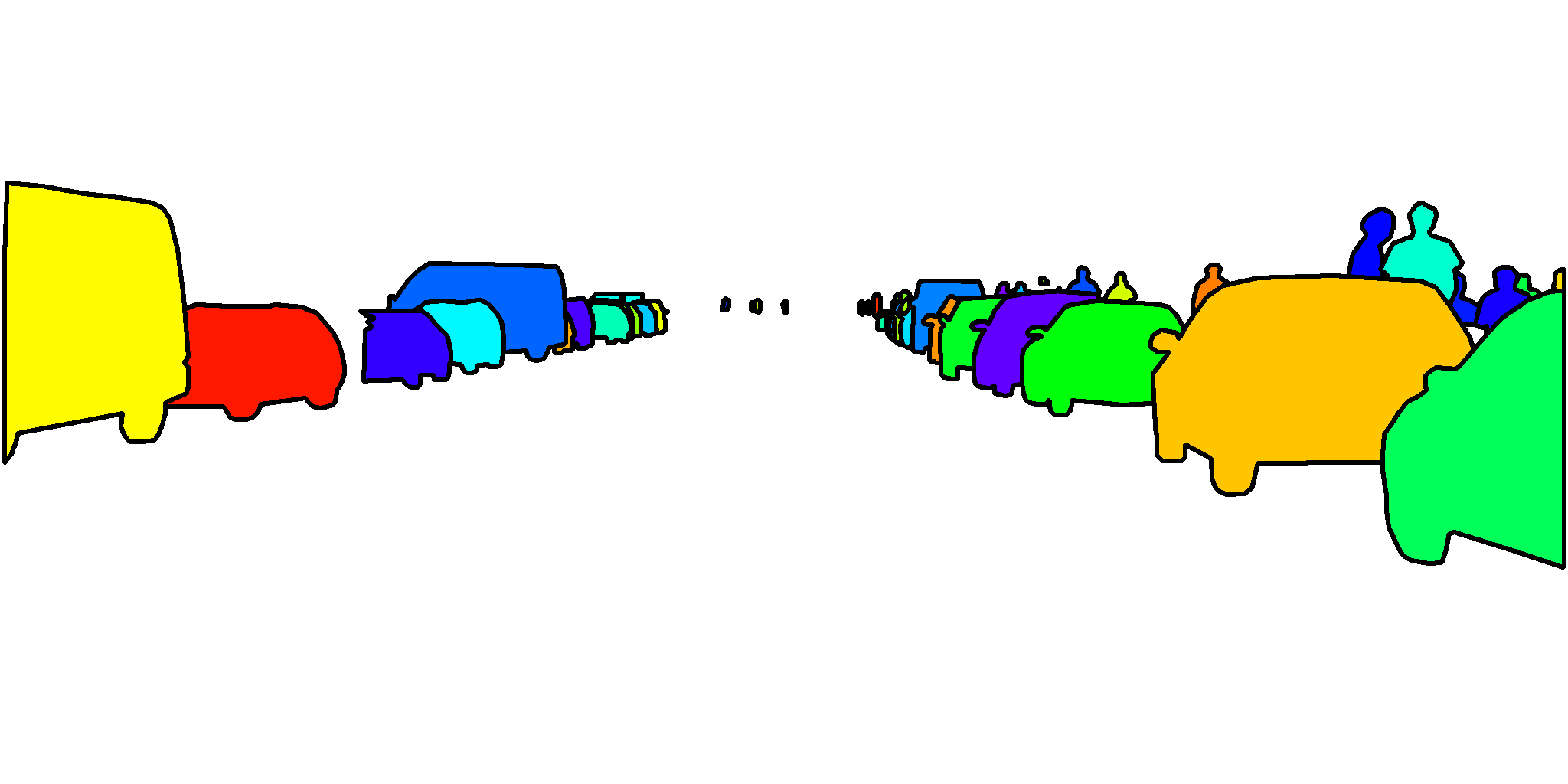}\end{subfigure}

\vspace{0.10cm}

\begin{subfigure}[b]{0.22\textwidth} \centering \includegraphics[trim={0 3cm 0 1.25cm},clip,width=\textwidth]{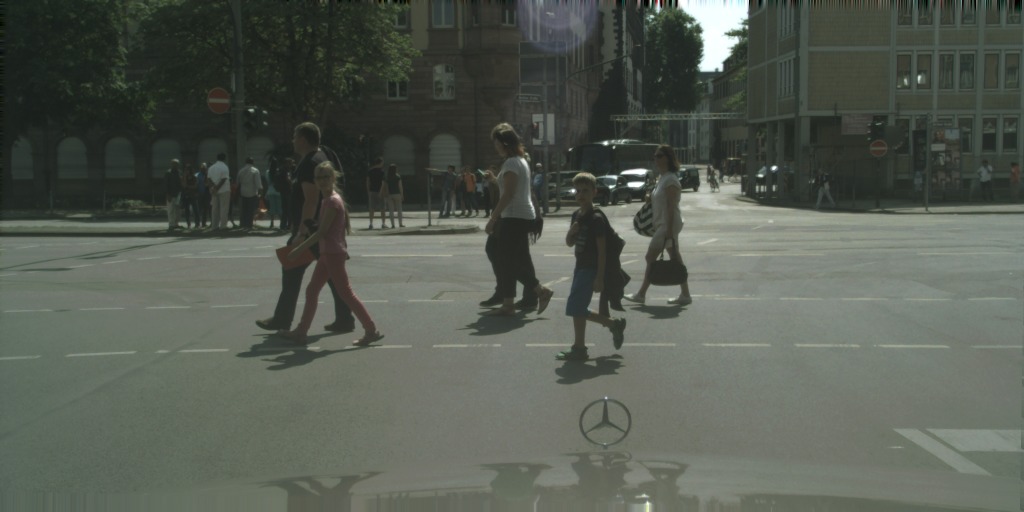}\end{subfigure}
\begin{subfigure}[b]{0.22\textwidth} \centering \includegraphics[trim={0 3cm 0 1.25cm},clip,width=\textwidth]{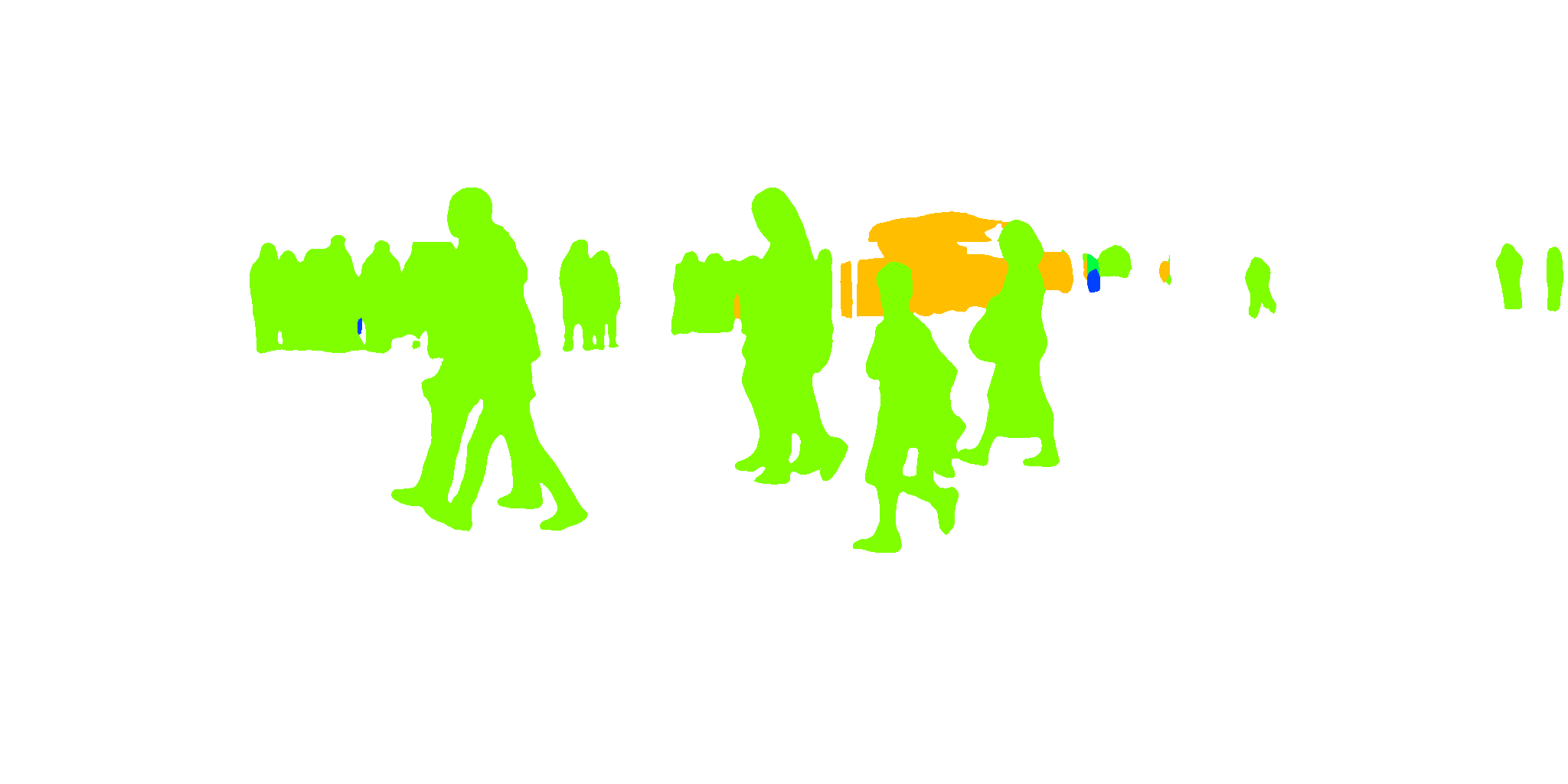}\end{subfigure}
\begin{subfigure}[b]{0.22\textwidth} \centering \includegraphics[trim={0 3cm 0 1.25cm},clip,width=\textwidth]{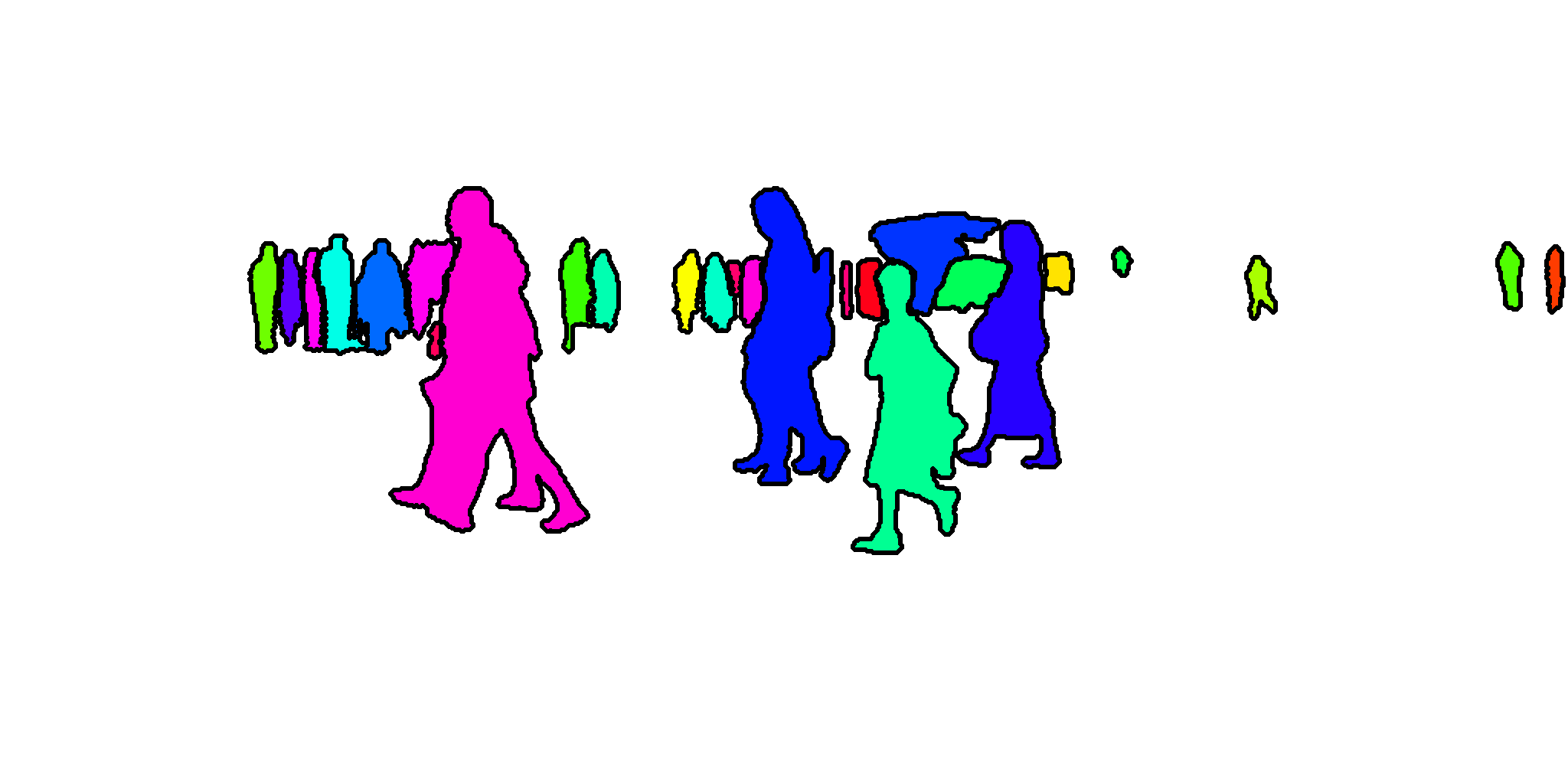}\end{subfigure}
\begin{subfigure}[b]{0.22\textwidth} \centering \includegraphics[trim={0 3cm 0 1.25cm},clip,width=\textwidth]{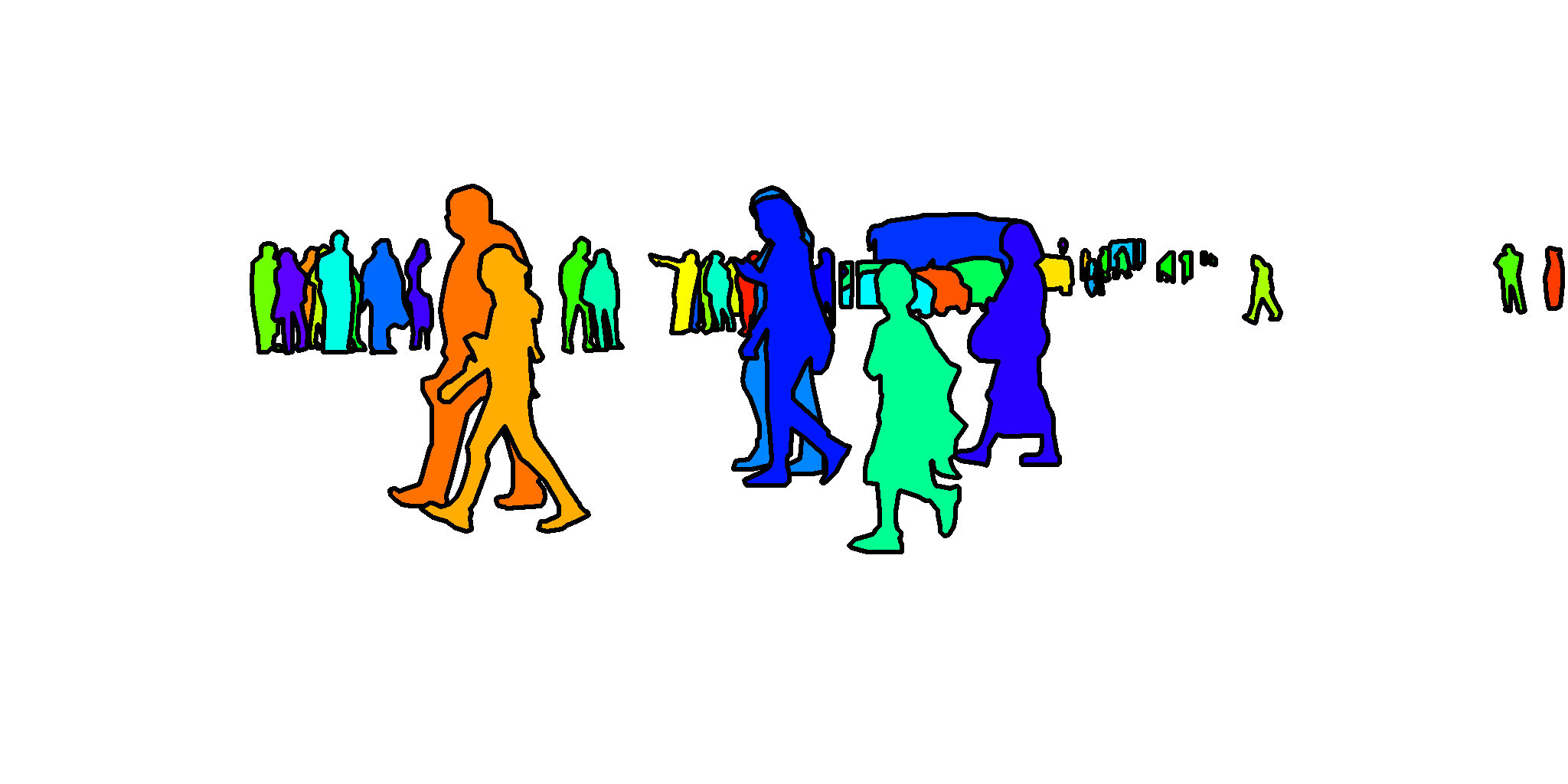}\end{subfigure}

\vspace{0.10cm}

\begin{subfigure}[b]{0.22\textwidth} \centering \includegraphics[trim={0 3cm 0 1.25cm},clip,width=\textwidth]{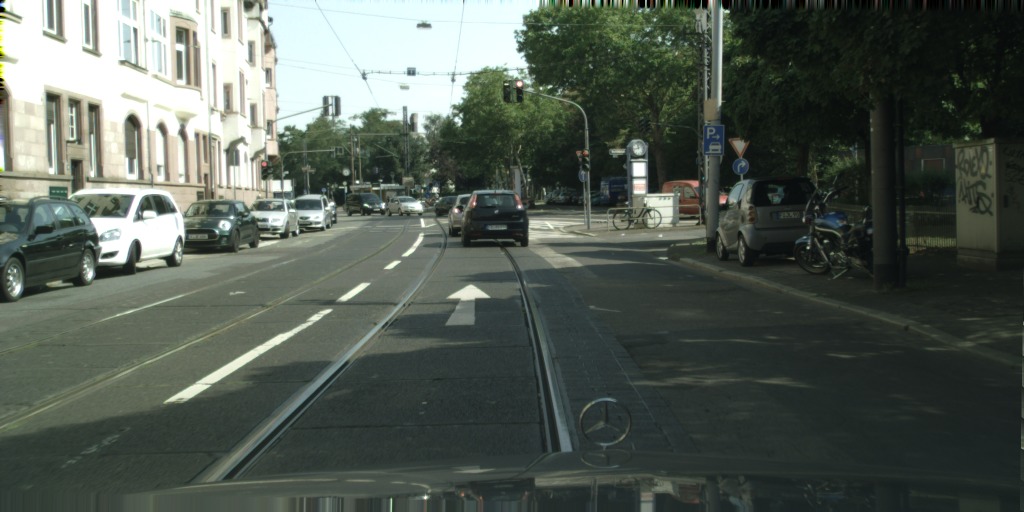}\end{subfigure}
\begin{subfigure}[b]{0.22\textwidth} \centering \includegraphics[trim={0 3cm 0 1.25cm},clip,width=\textwidth]{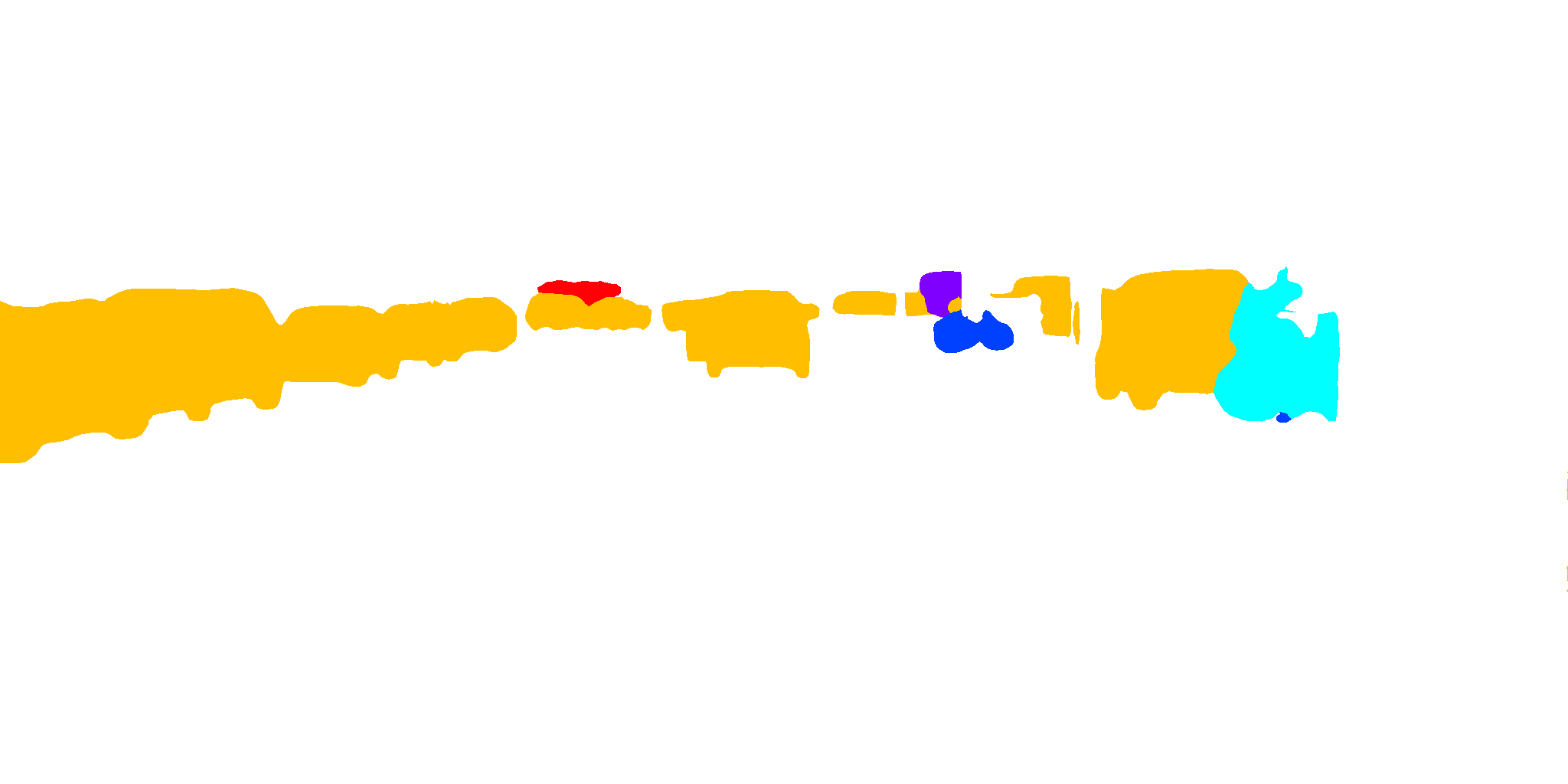}\end{subfigure}
\begin{subfigure}[b]{0.22\textwidth} \centering \includegraphics[trim={0 3cm 0 1.25cm},clip,width=\textwidth]{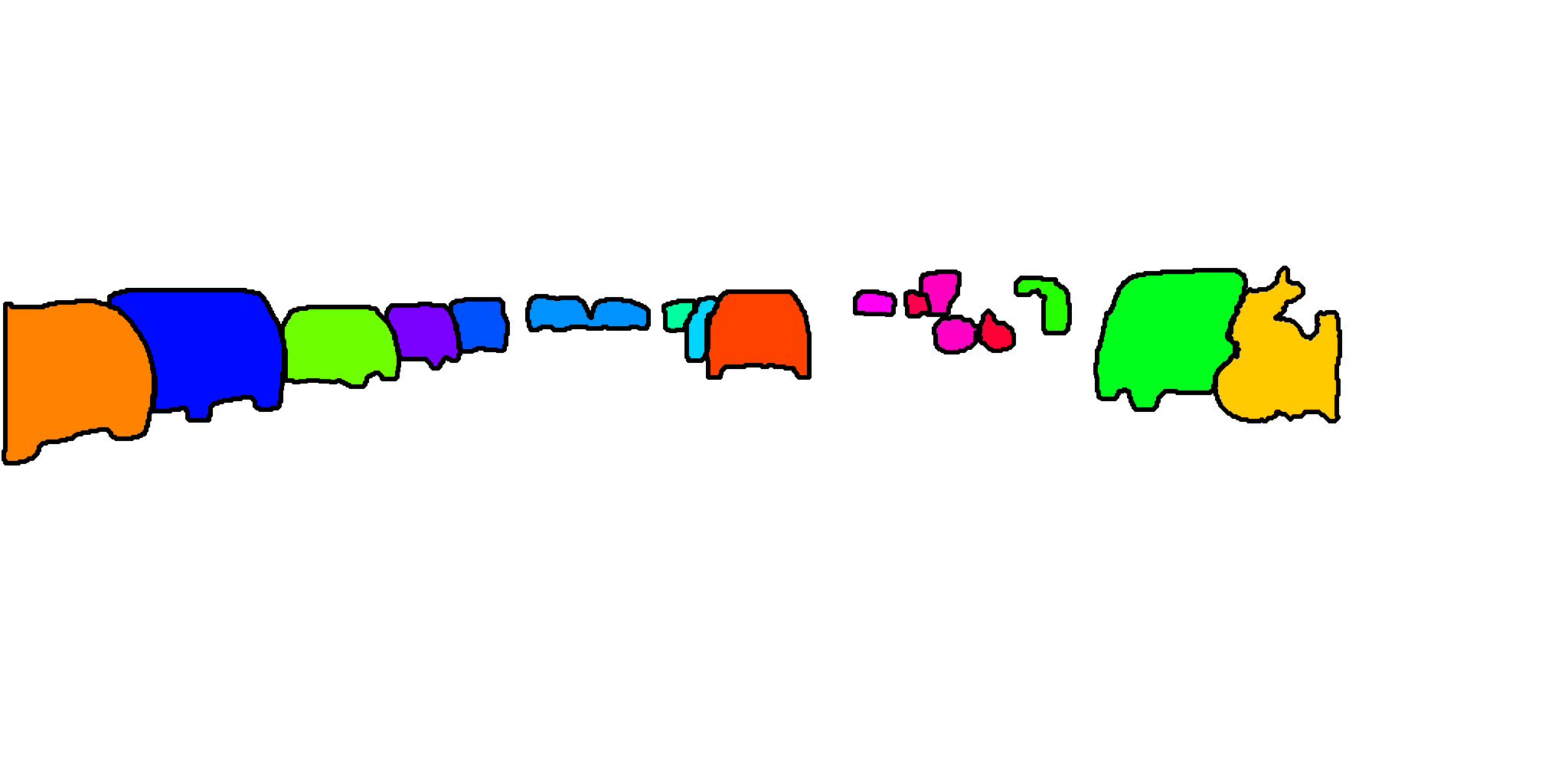}\end{subfigure}
\begin{subfigure}[b]{0.22\textwidth} \centering \includegraphics[trim={0 3cm 0 1.25cm},clip,width=\textwidth]{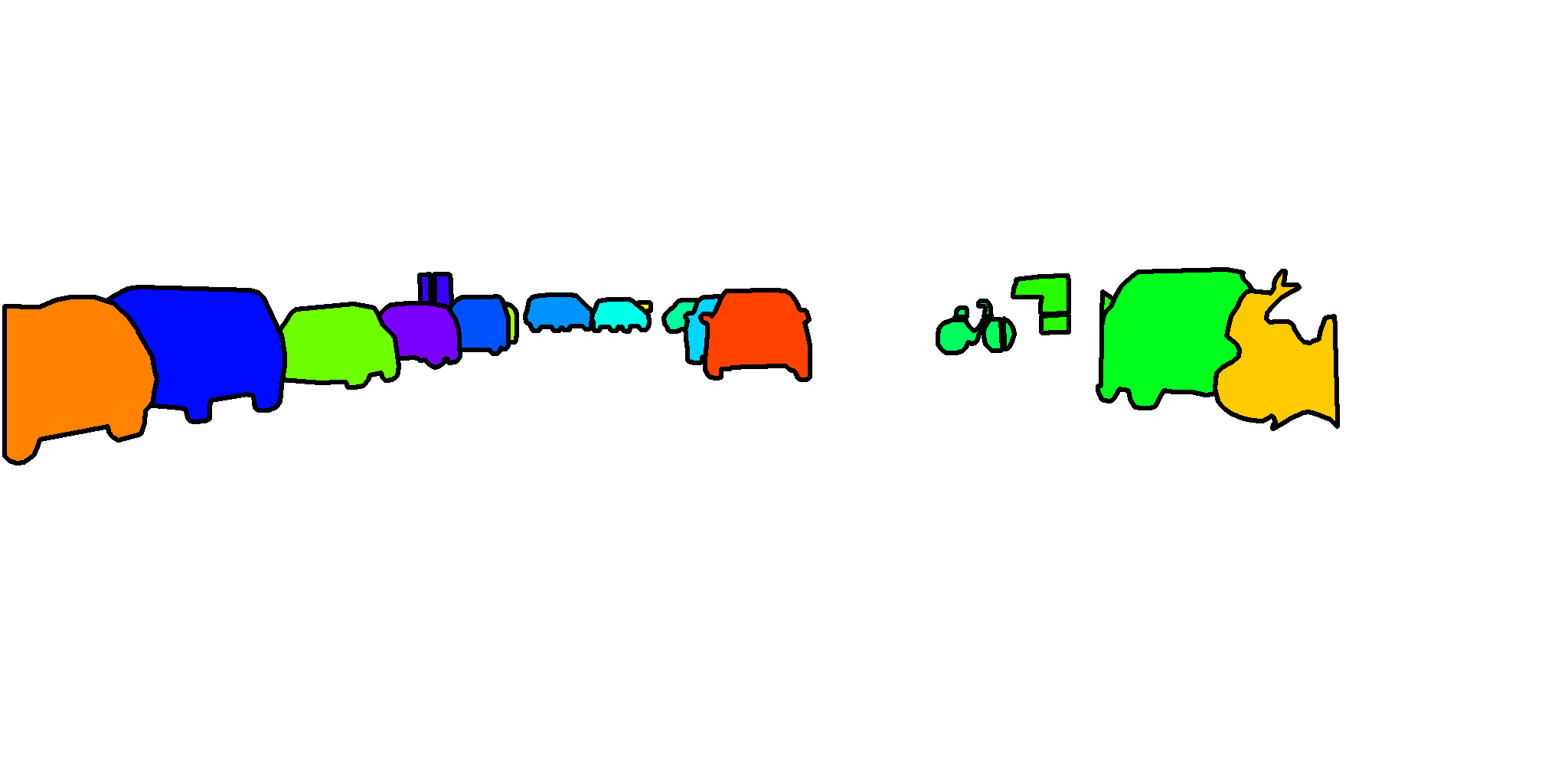}\end{subfigure}

\vspace{0.10cm}

\begin{subfigure}[b]{0.22\textwidth} \centering \includegraphics[trim={0 3cm 0 1.25cm},clip,width=\textwidth]{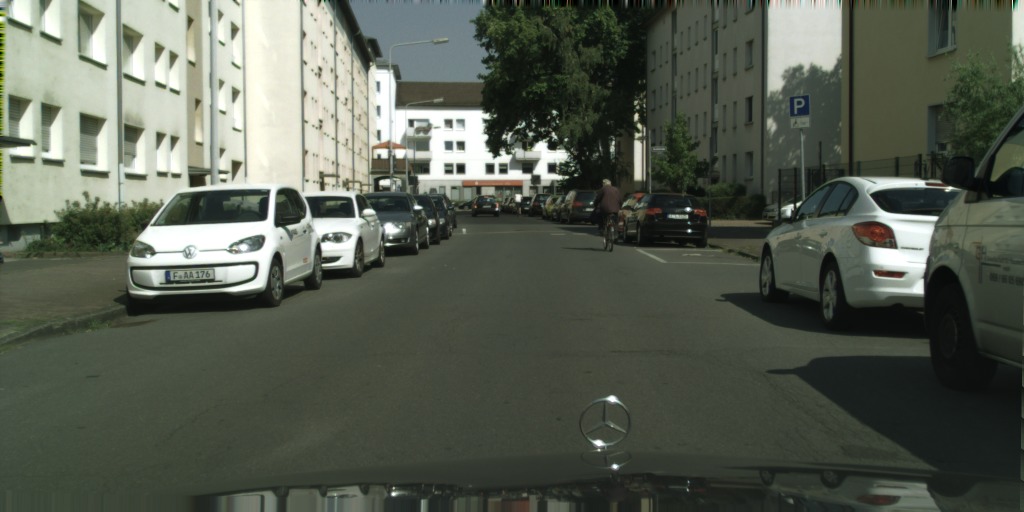}\end{subfigure}
\begin{subfigure}[b]{0.22\textwidth} \centering \includegraphics[trim={0 3cm 0 1.25cm},clip,width=\textwidth]{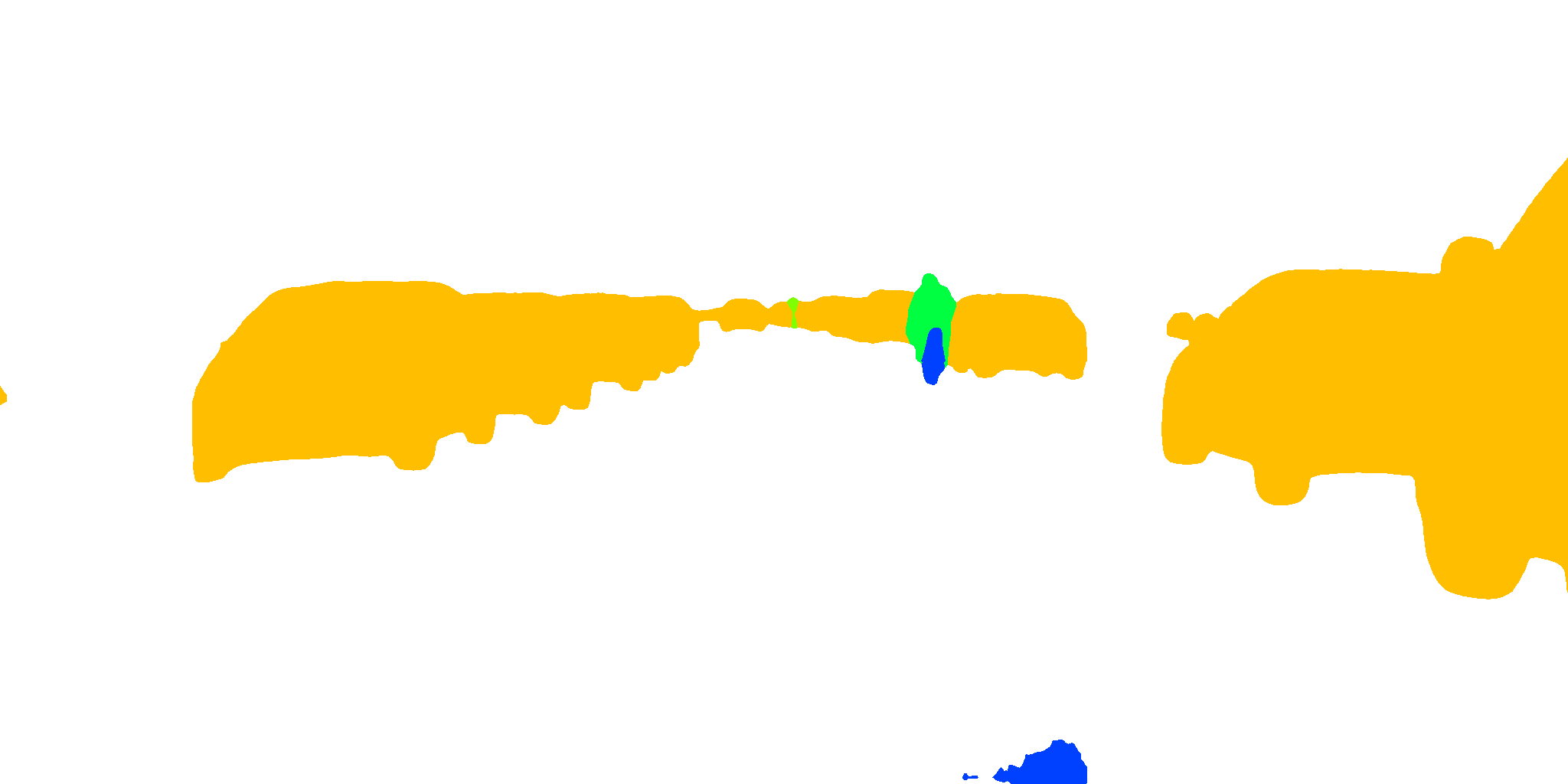}\end{subfigure}
\begin{subfigure}[b]{0.22\textwidth} \centering \includegraphics[trim={0 3cm 0 1.25cm},clip,width=\textwidth]{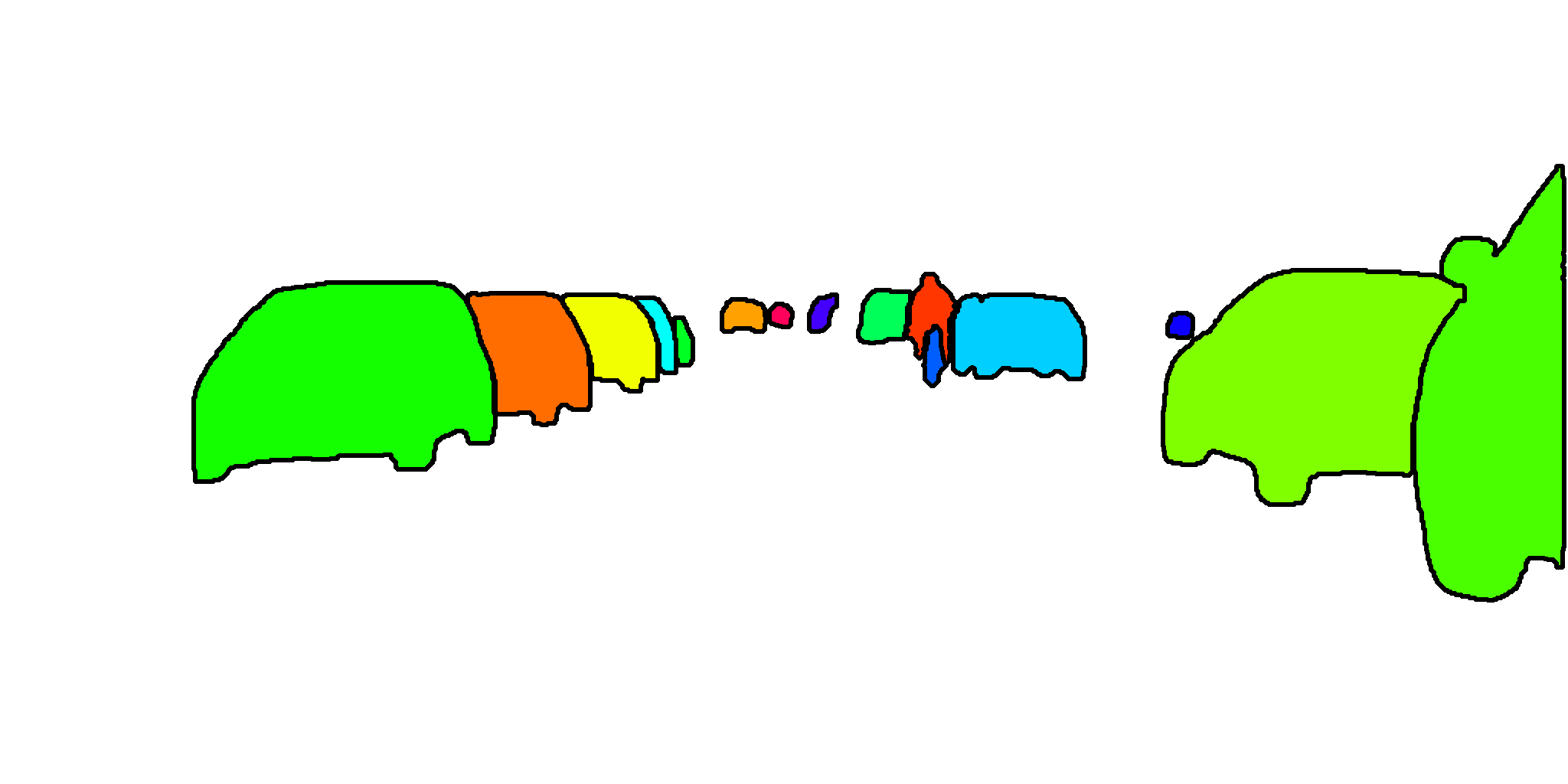}\end{subfigure}
\begin{subfigure}[b]{0.22\textwidth} \centering \includegraphics[trim={0 3cm 0 1.25cm},clip,width=\textwidth]{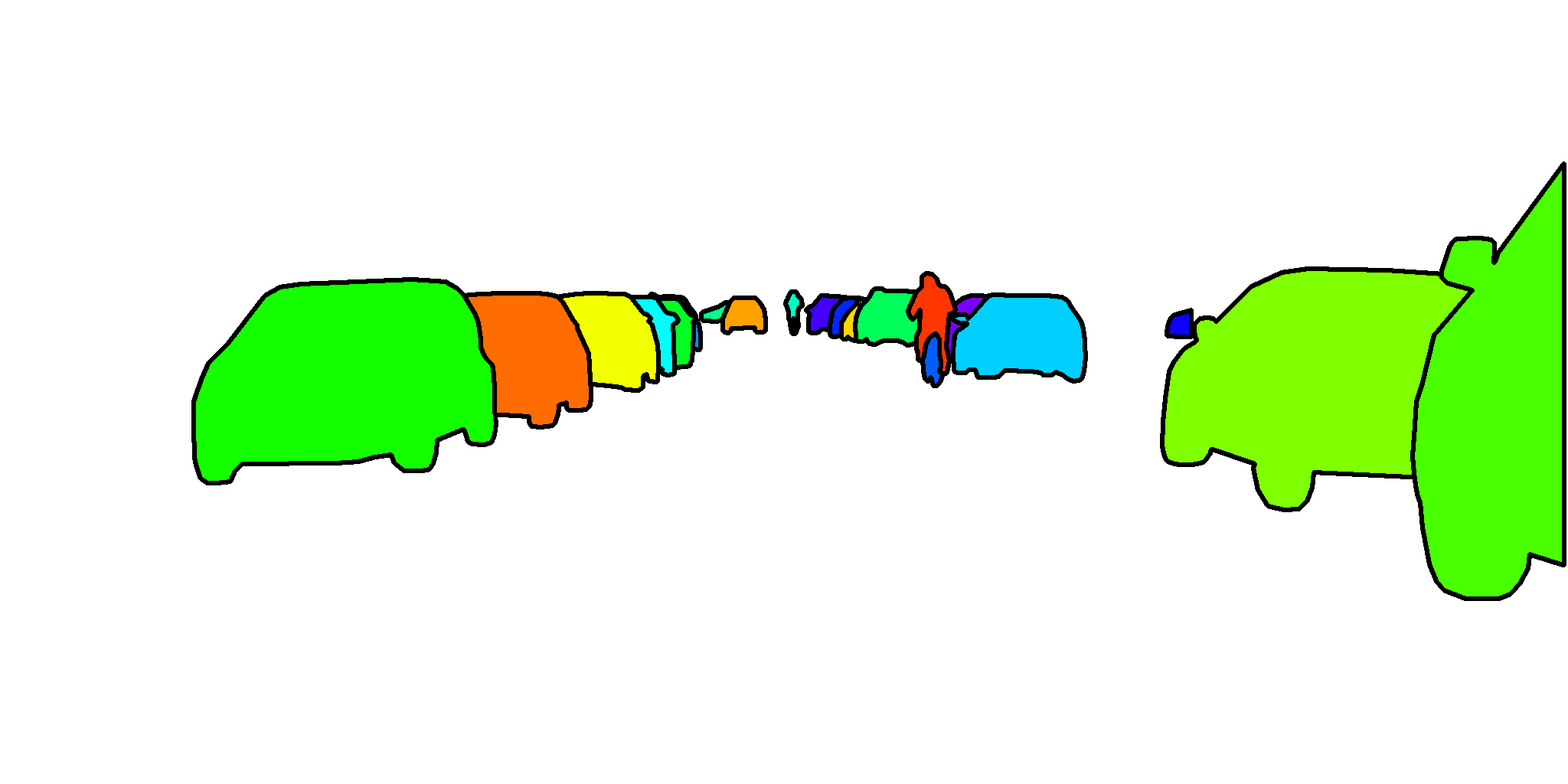}\end{subfigure}

\vspace{0.10cm}

\begin{subfigure}[b]{0.22\textwidth} \centering \includegraphics[trim={0 3cm 0 1.25cm},clip,width=\textwidth]{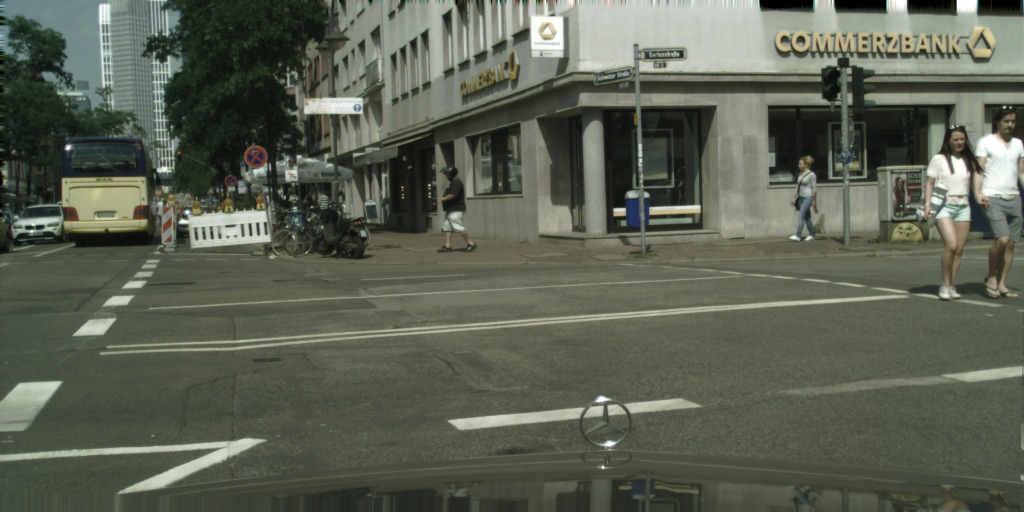}\end{subfigure}
\begin{subfigure}[b]{0.22\textwidth} \centering \includegraphics[trim={0 3cm 0 1.25cm},clip,width=\textwidth]{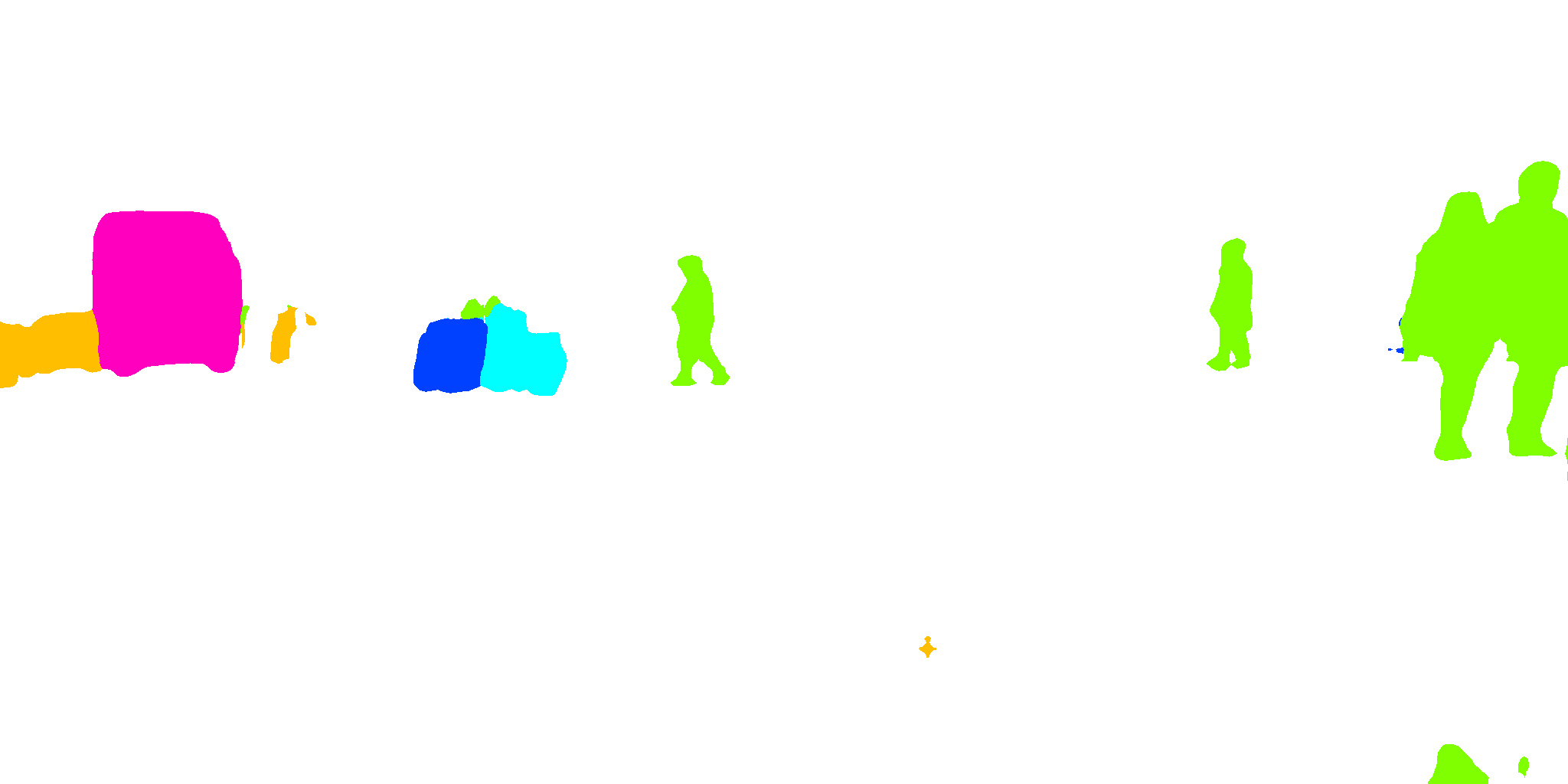}\end{subfigure}
\begin{subfigure}[b]{0.22\textwidth} \centering \includegraphics[trim={0 3cm 0 1.25cm},clip,width=\textwidth]{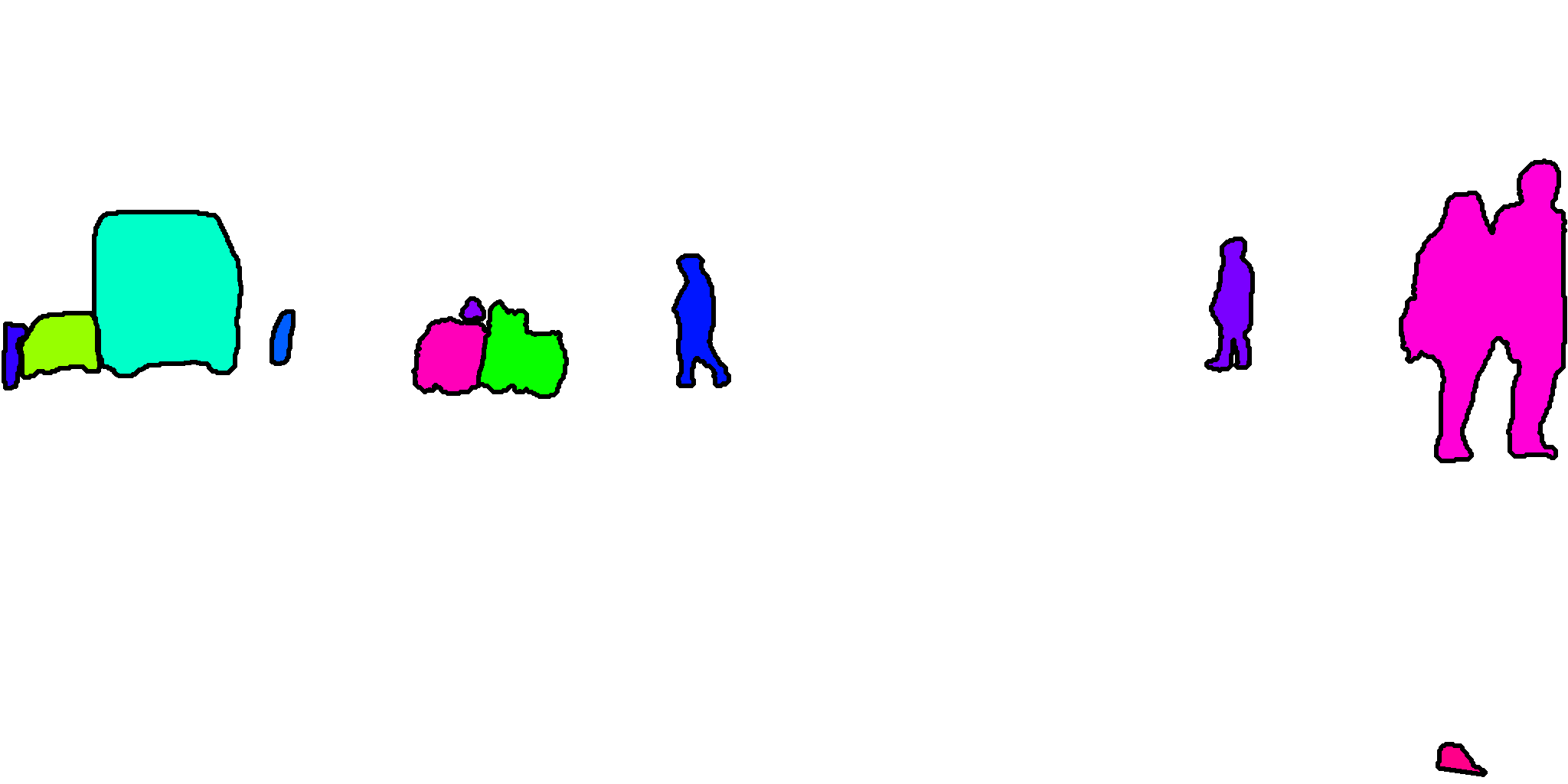}\end{subfigure}
\begin{subfigure}[b]{0.22\textwidth} \centering \includegraphics[trim={0 3cm 0 1.25cm},clip,width=\textwidth]{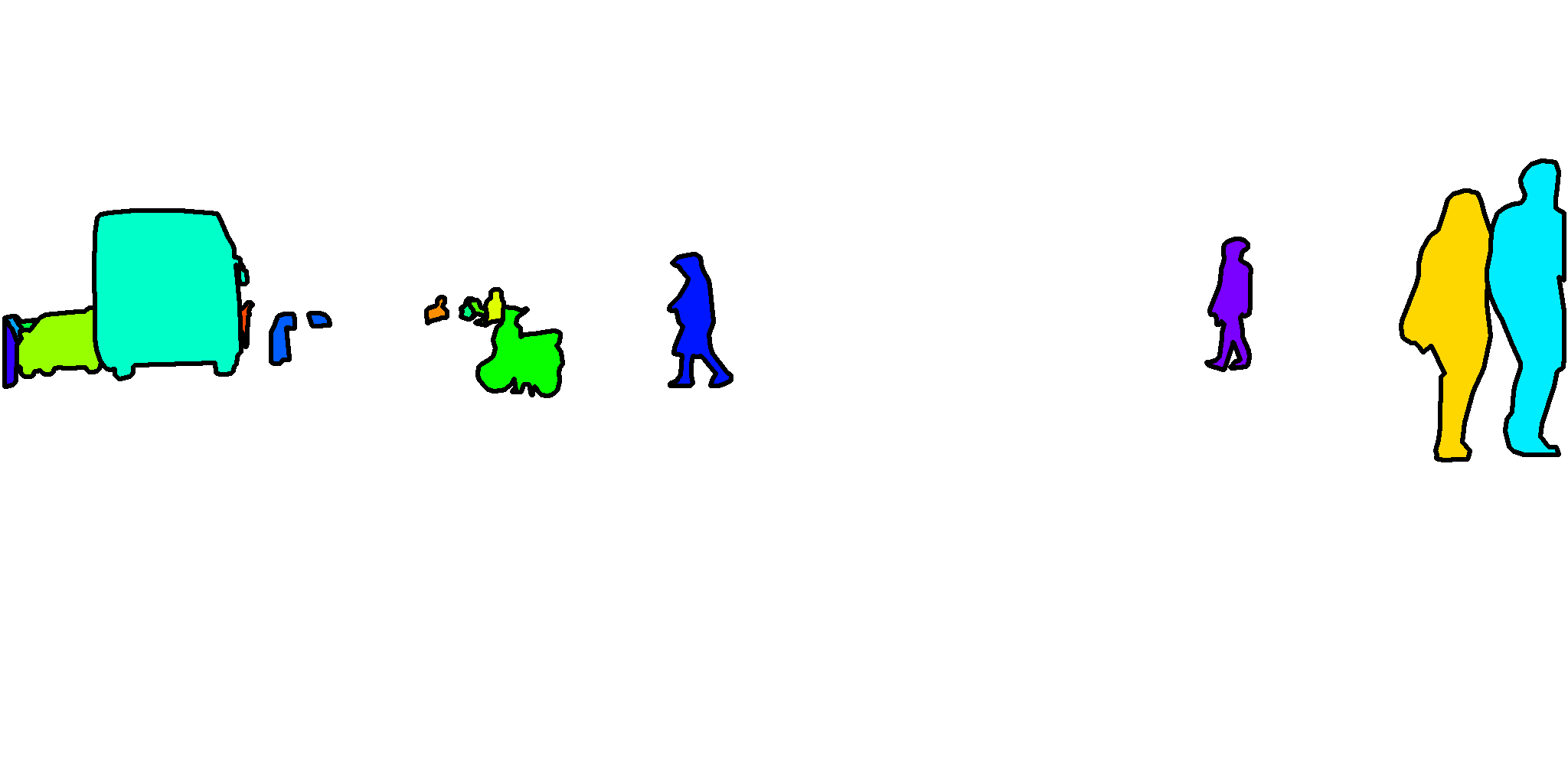}\end{subfigure}

\vspace{0.10cm}

\begin{subfigure}[b]{0.22\textwidth} \centering \includegraphics[trim={0 3cm 0 1.25cm},clip,width=\textwidth]{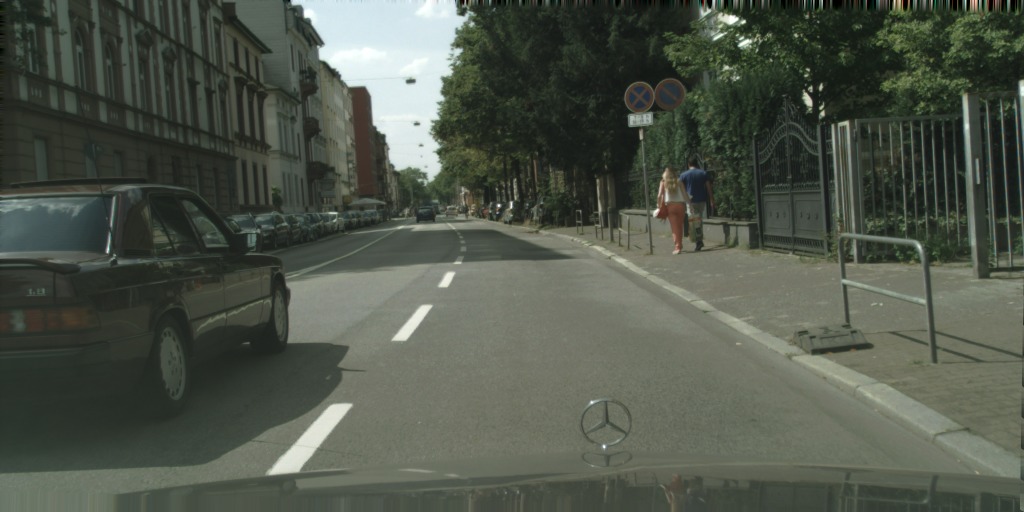}\end{subfigure}
\begin{subfigure}[b]{0.22\textwidth} \centering \includegraphics[trim={0 3cm 0 1.25cm},clip,width=\textwidth]{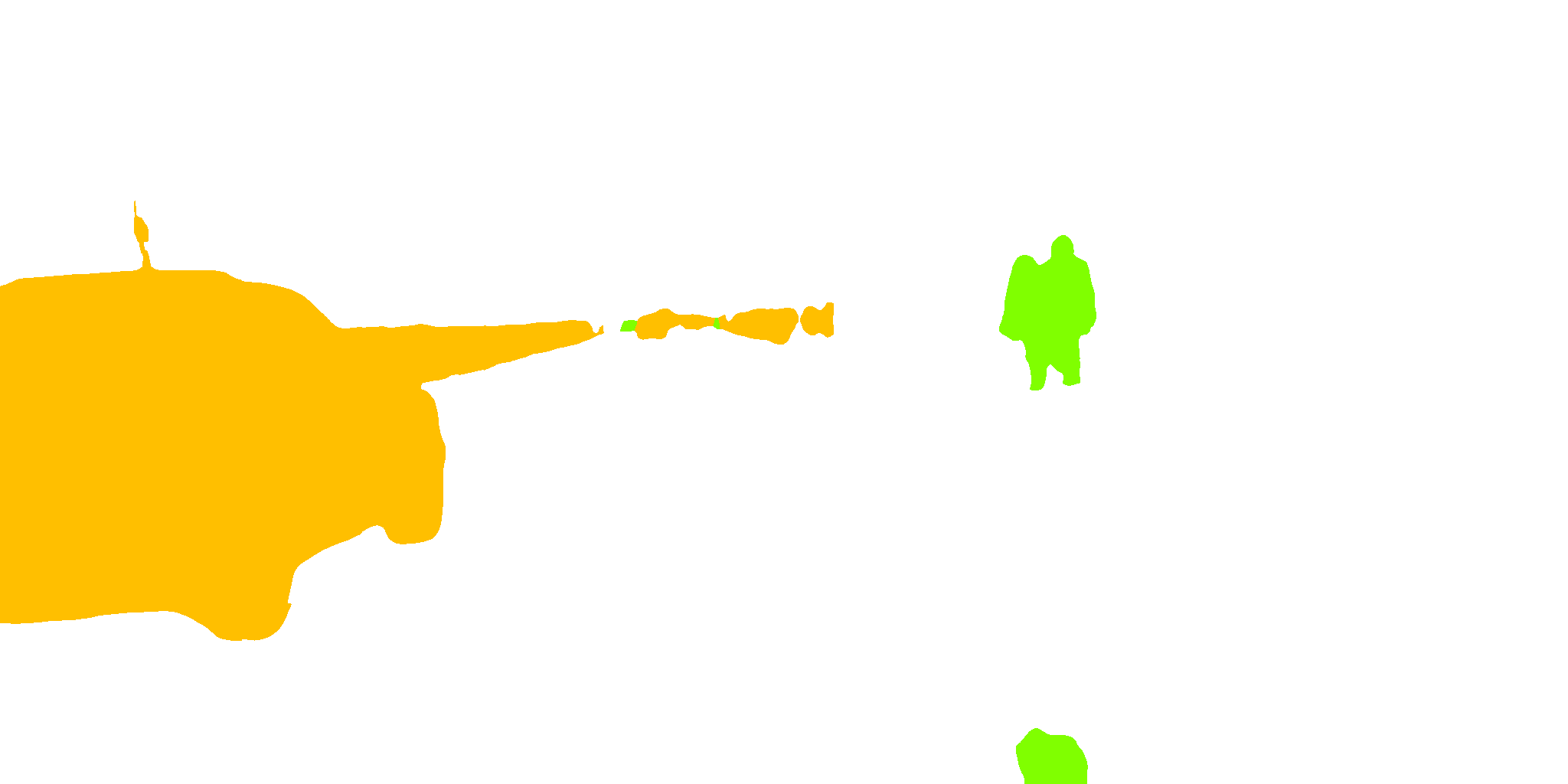}\end{subfigure}
\begin{subfigure}[b]{0.22\textwidth} \centering \includegraphics[trim={0 3cm 0 1.25cm},clip,width=\textwidth]{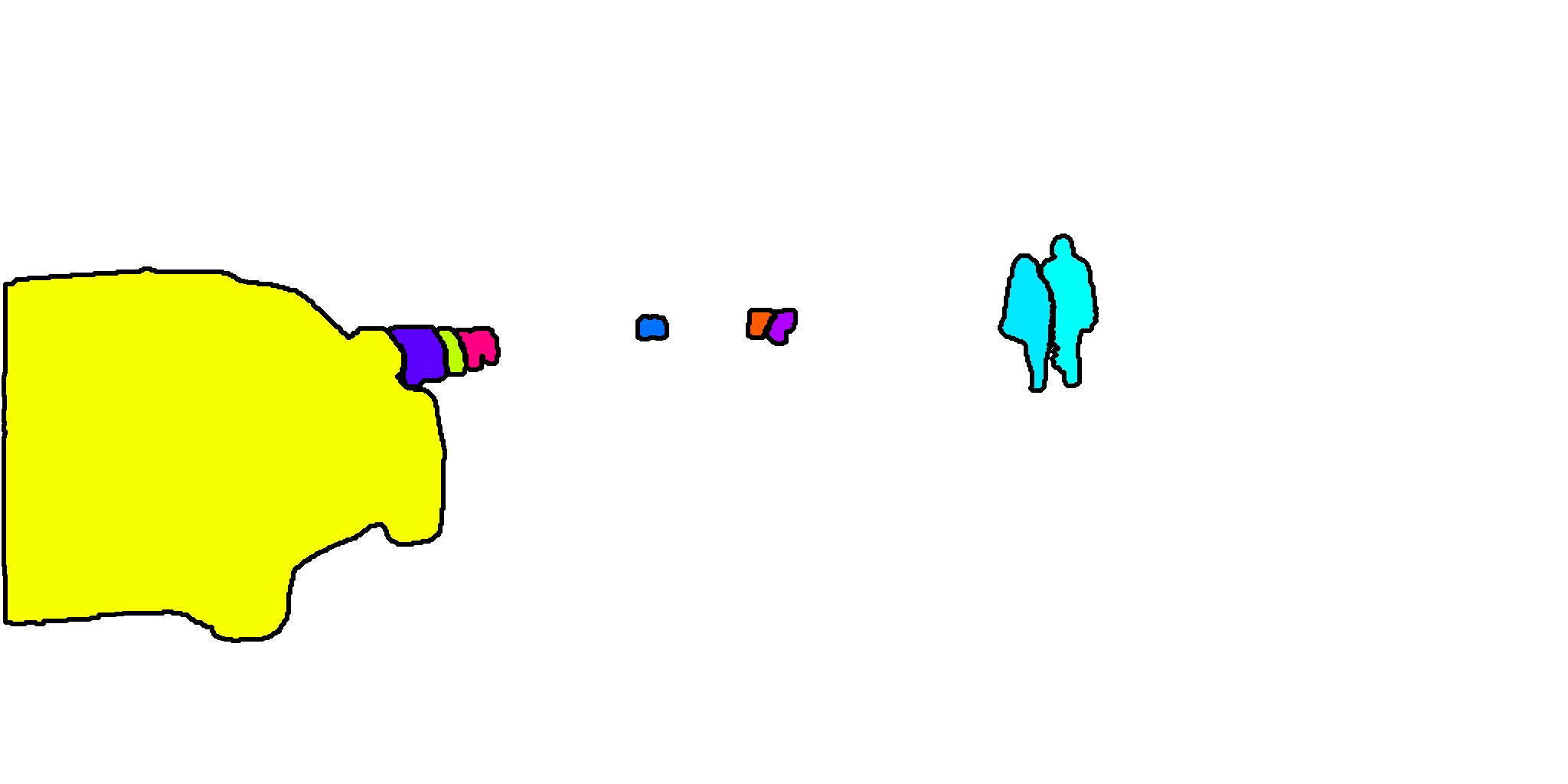}\end{subfigure}
\begin{subfigure}[b]{0.22\textwidth} \centering \includegraphics[trim={0 3cm 0 1.25cm},clip,width=\textwidth]{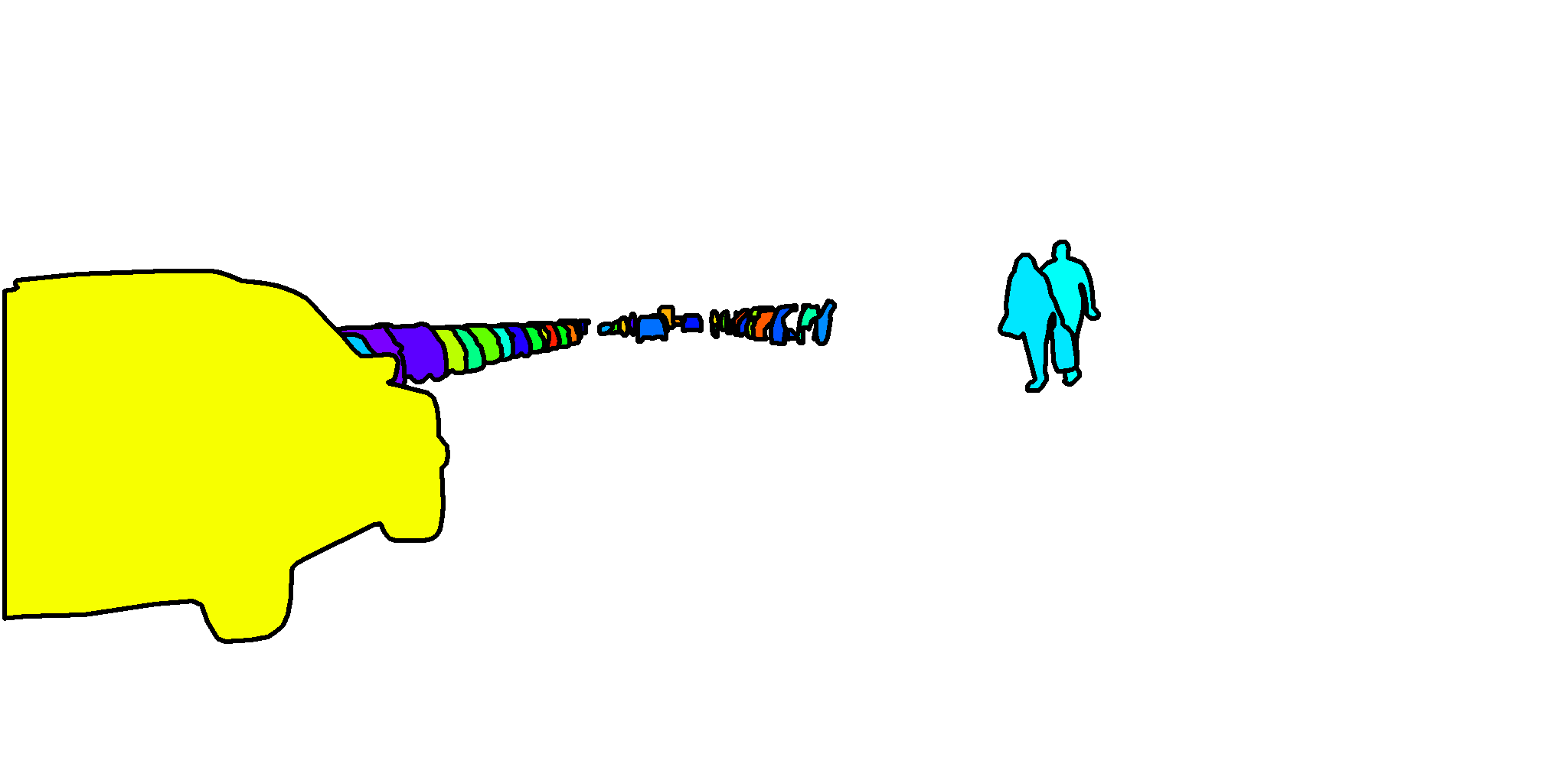}\end{subfigure}

\vspace{0.10cm}

\begin{subfigure}[b]{0.22\textwidth} \centering \includegraphics[trim={0 3cm 0 1.25cm},clip,width=\textwidth]{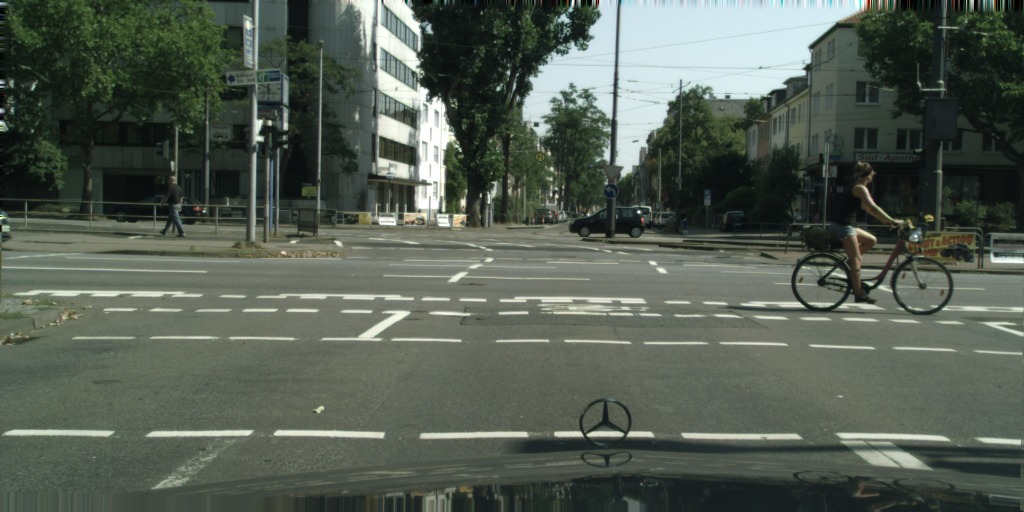}\end{subfigure}
\begin{subfigure}[b]{0.22\textwidth} \centering \includegraphics[trim={0 3cm 0 1.25cm},clip,width=\textwidth]{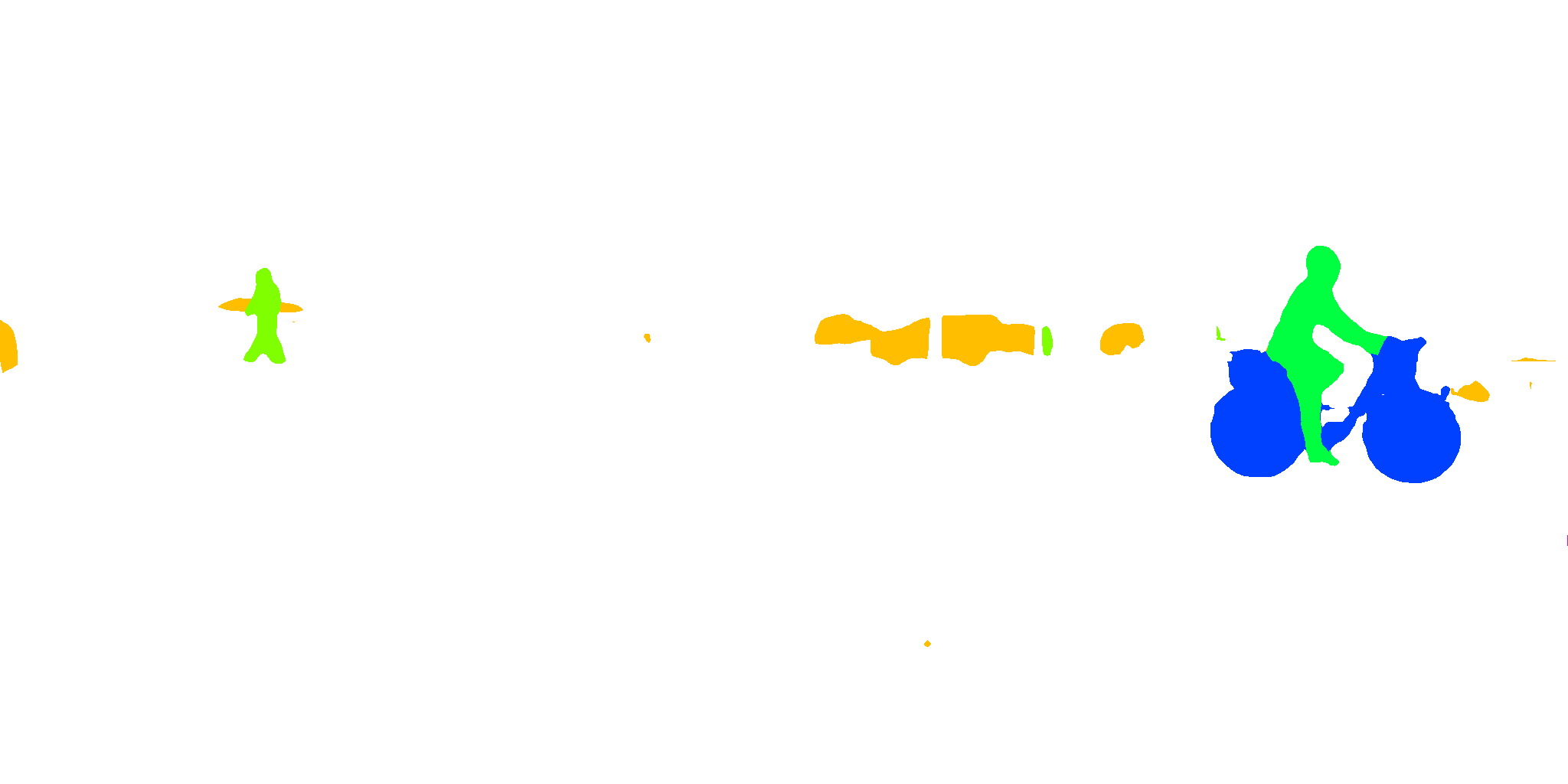}\end{subfigure}
\begin{subfigure}[b]{0.22\textwidth} \centering \includegraphics[trim={0 3cm 0 1.25cm},clip,width=\textwidth]{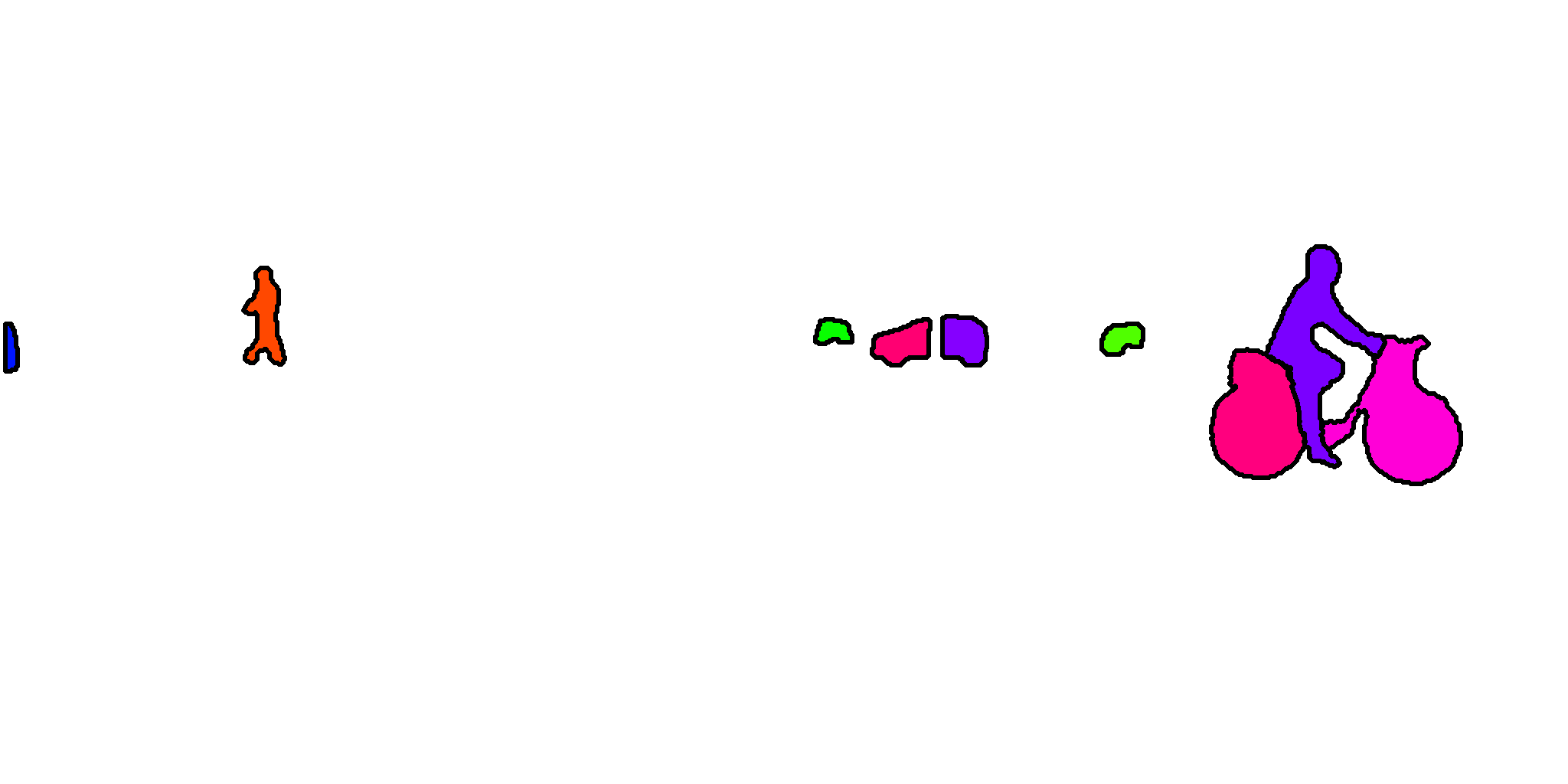}\end{subfigure}
\begin{subfigure}[b]{0.22\textwidth} \centering \includegraphics[trim={0 3cm 0 1.25cm},clip,width=\textwidth]{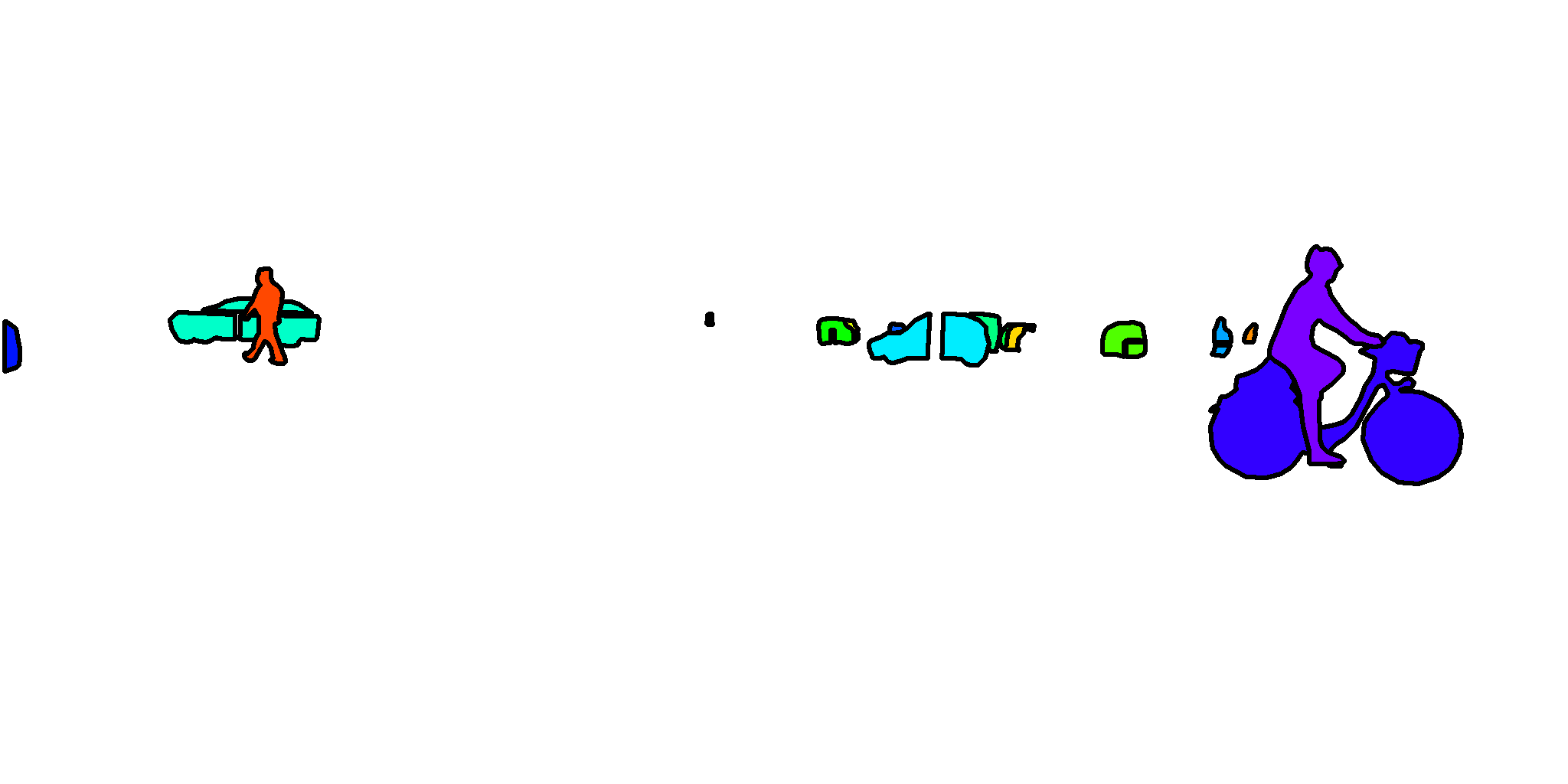}\end{subfigure}

\vspace{0.10cm}

\begin{subfigure}[b]{0.22\textwidth} \centering \includegraphics[trim={0 3cm 0 1.25cm},clip,width=\textwidth]{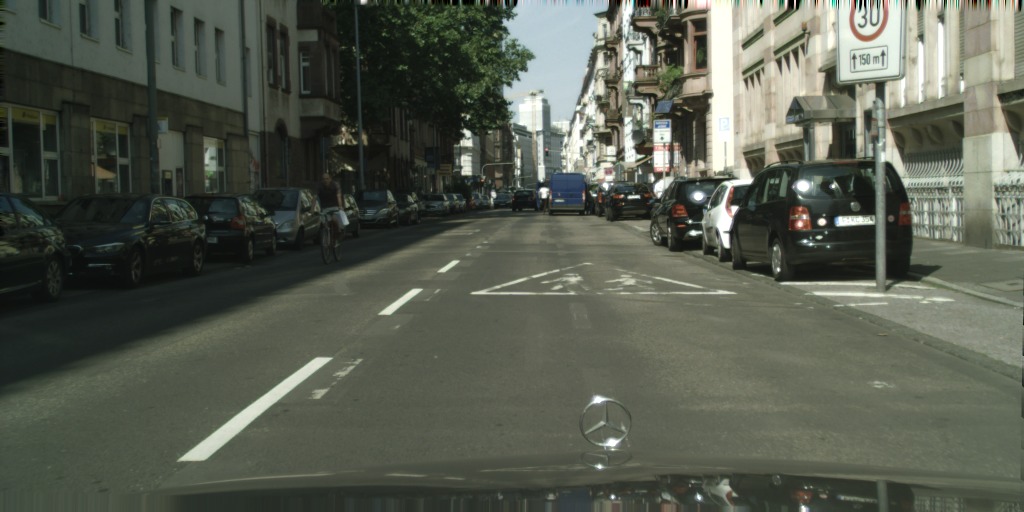}\end{subfigure}
\begin{subfigure}[b]{0.22\textwidth} \centering \includegraphics[trim={0 3cm 0 1.25cm},clip,width=\textwidth]{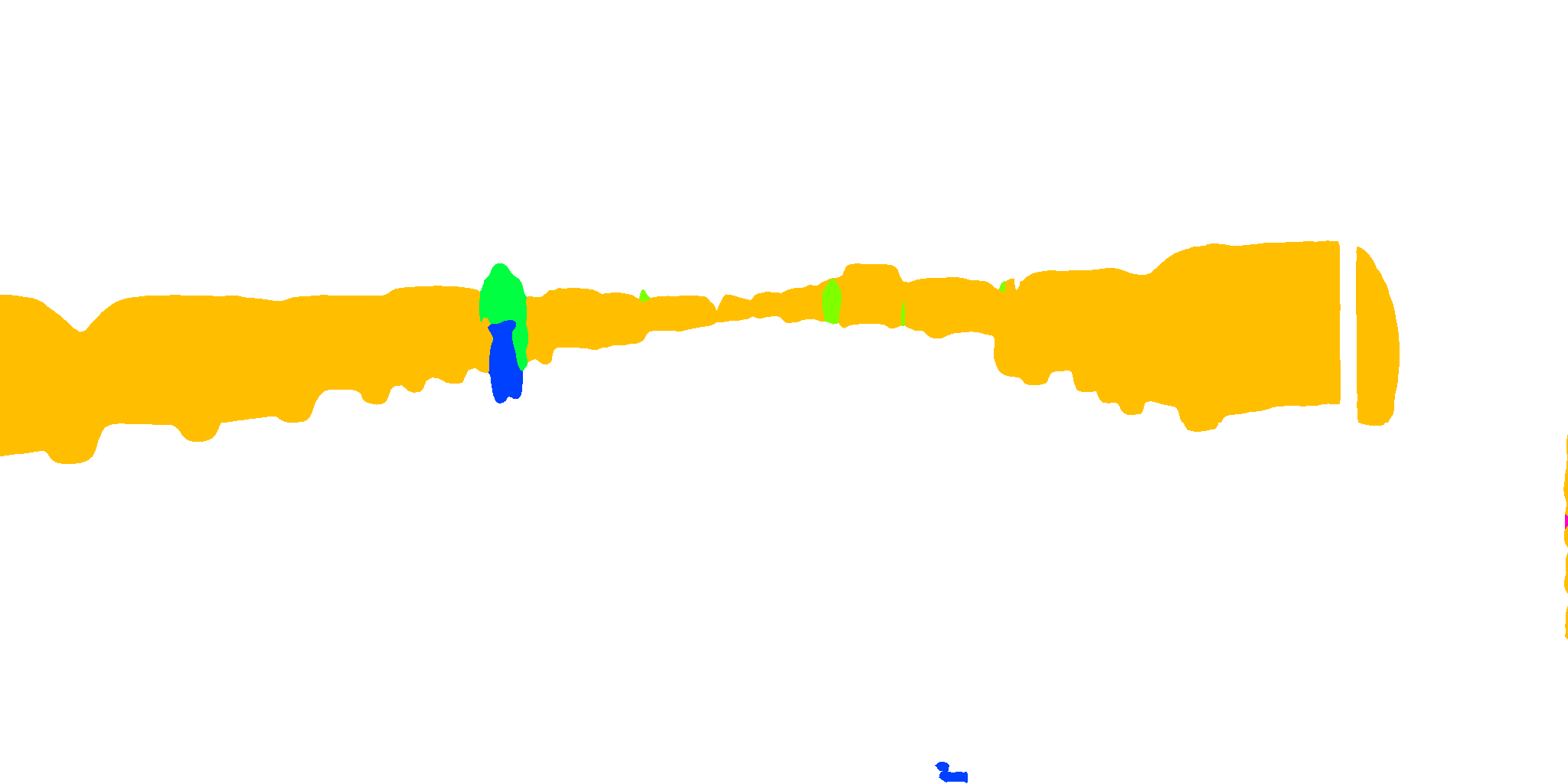}\end{subfigure}
\begin{subfigure}[b]{0.22\textwidth} \centering \includegraphics[trim={0 3cm 0 1.25cm},clip,width=\textwidth]{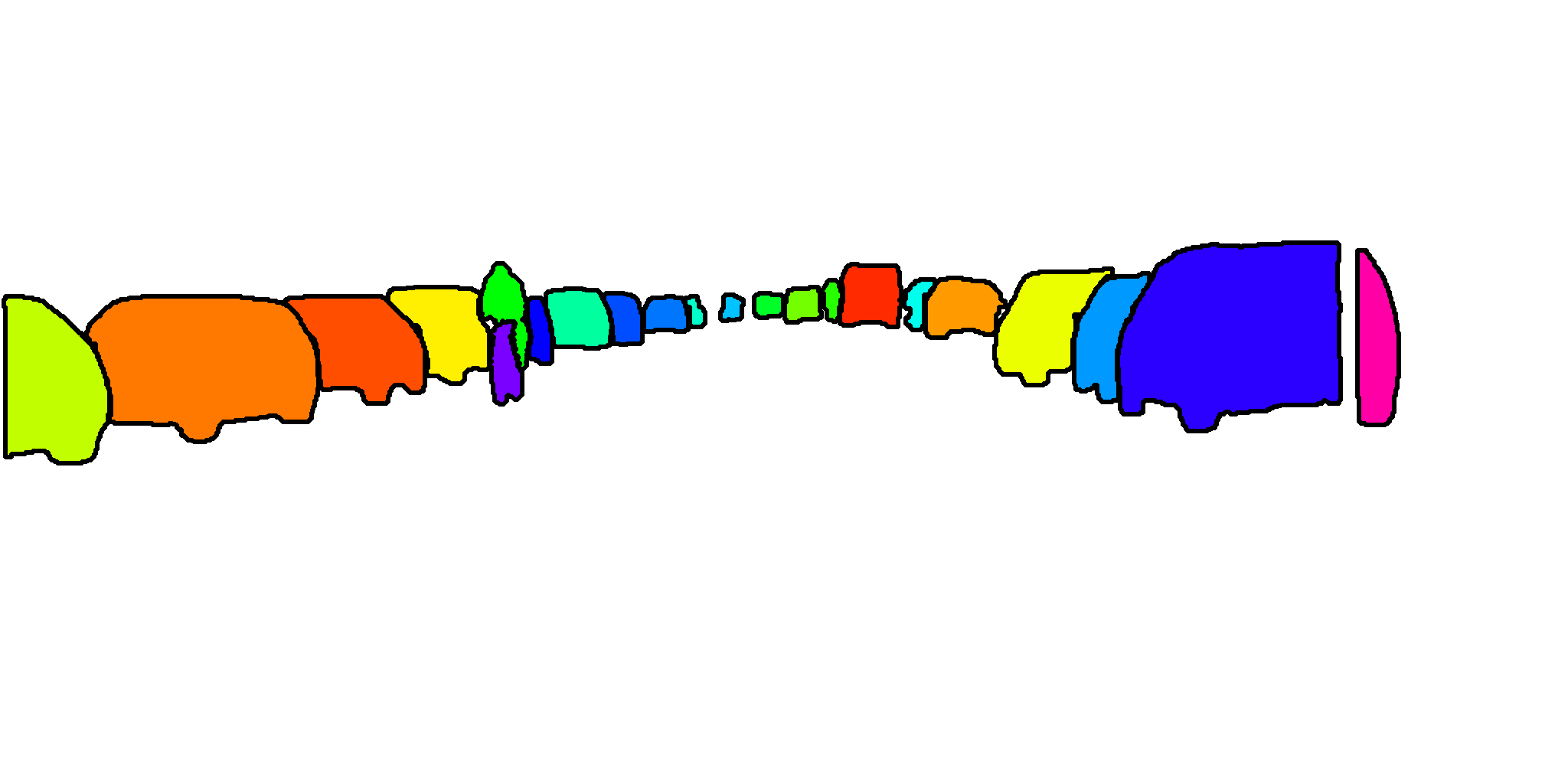}\end{subfigure}
\begin{subfigure}[b]{0.22\textwidth} \centering \includegraphics[trim={0 3cm 0 1.25cm},clip,width=\textwidth]{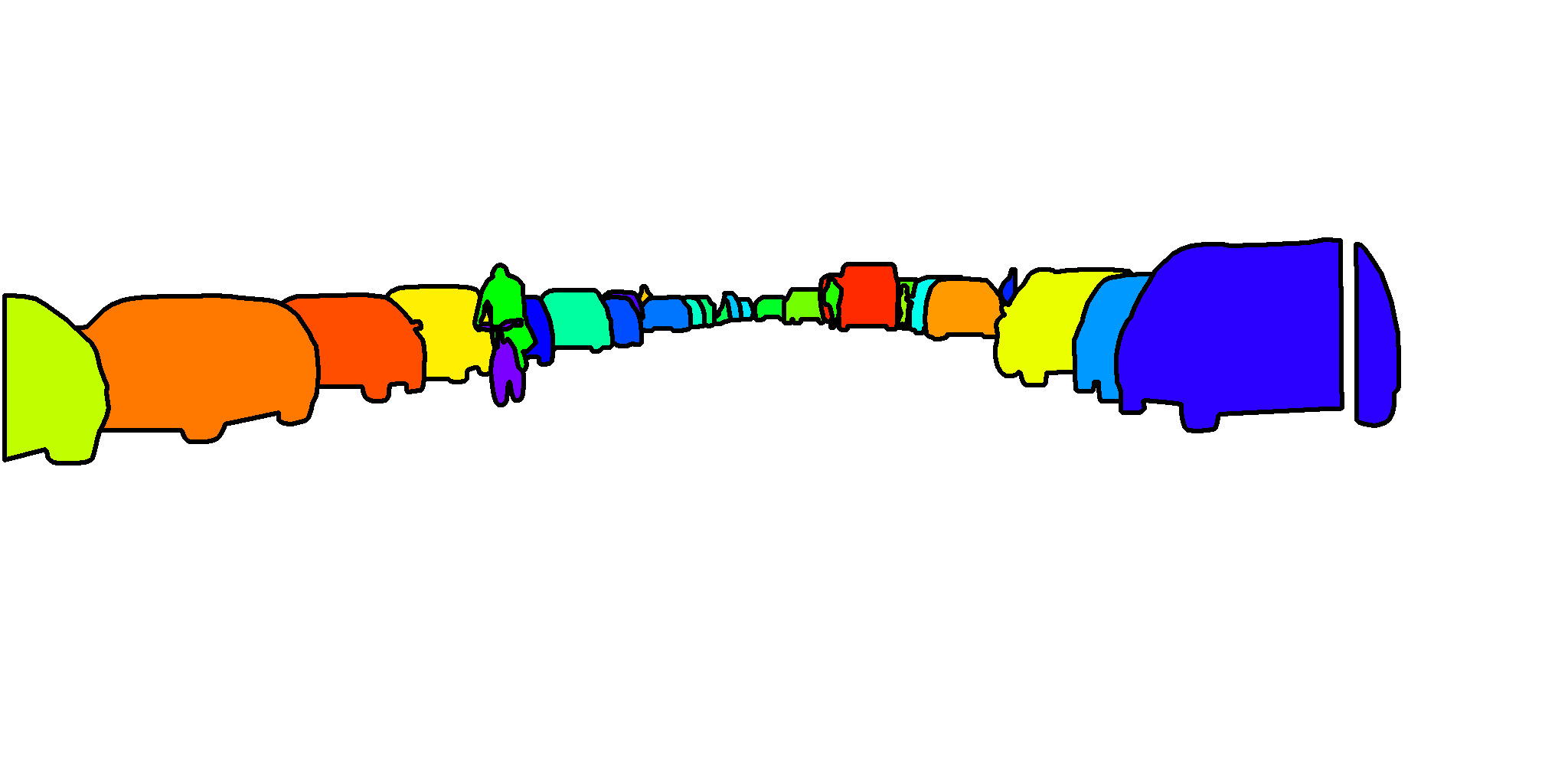}\end{subfigure}

\vspace{0.10cm}

\begin{subfigure}[b]{0.22\textwidth} \centering \includegraphics[trim={0 3cm 0 1.25cm},clip,width=\textwidth]{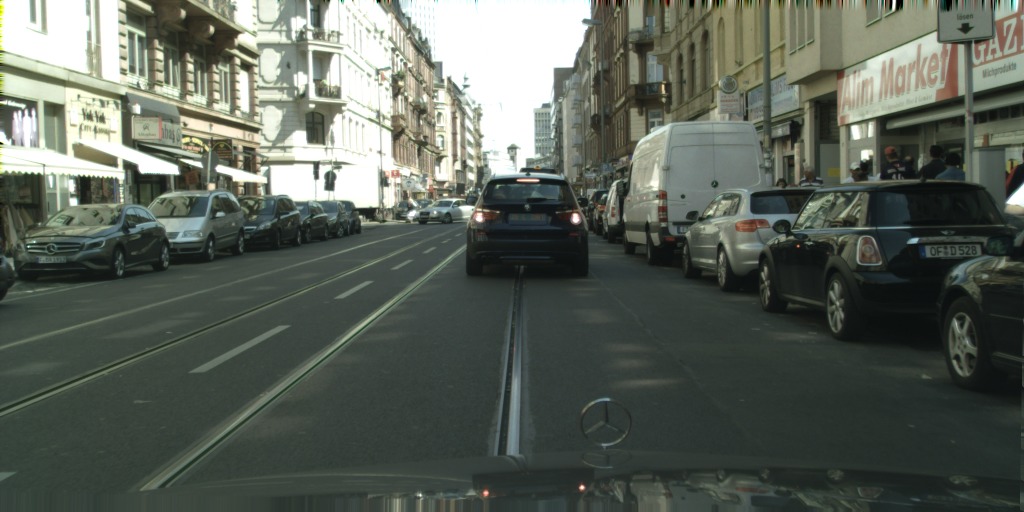}\end{subfigure}
\begin{subfigure}[b]{0.22\textwidth} \centering \includegraphics[trim={0 3cm 0 1.25cm},clip,width=\textwidth]{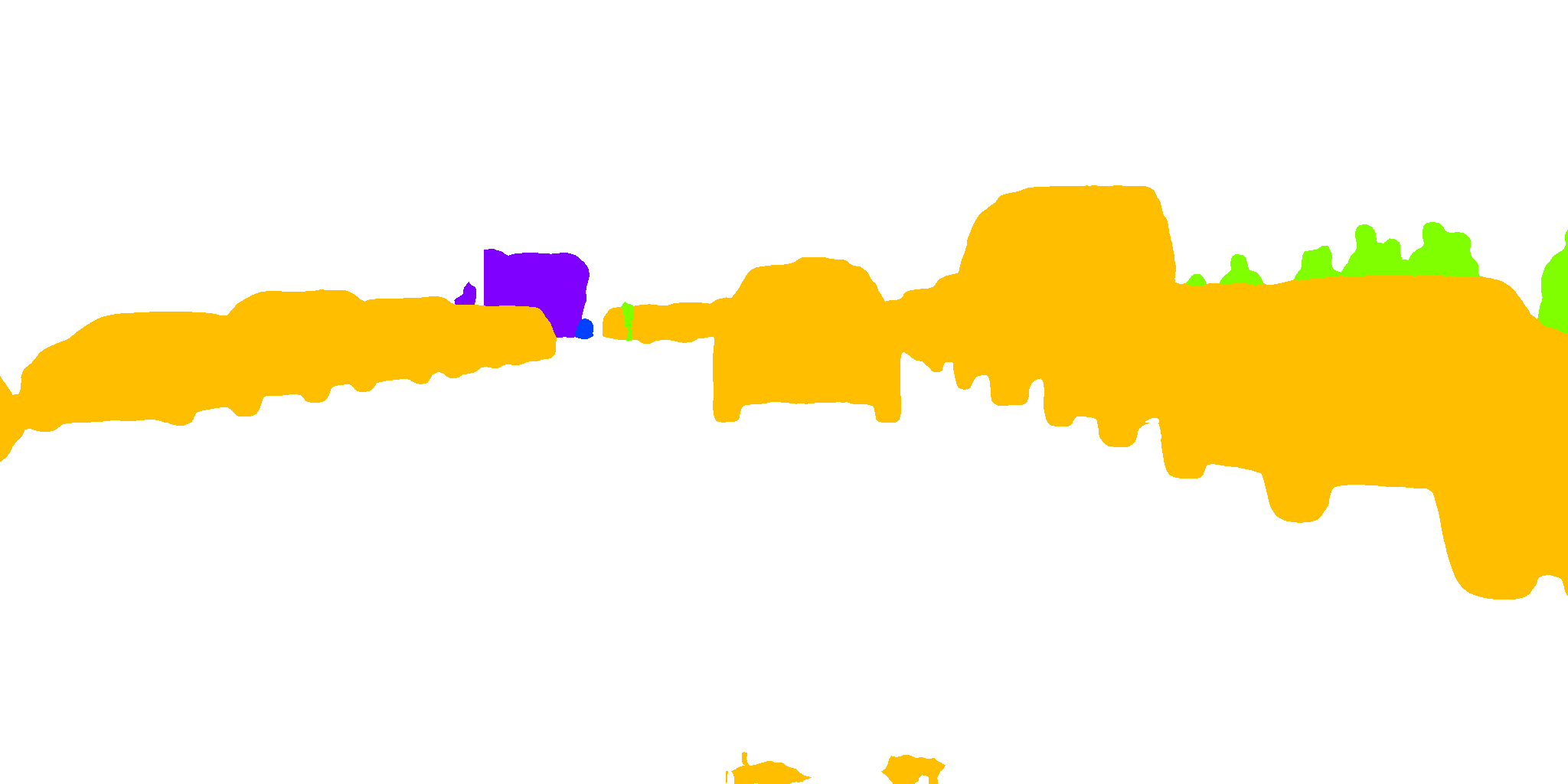}\end{subfigure}
\begin{subfigure}[b]{0.22\textwidth} \centering \includegraphics[trim={0 3cm 0 1.25cm},clip,width=\textwidth]{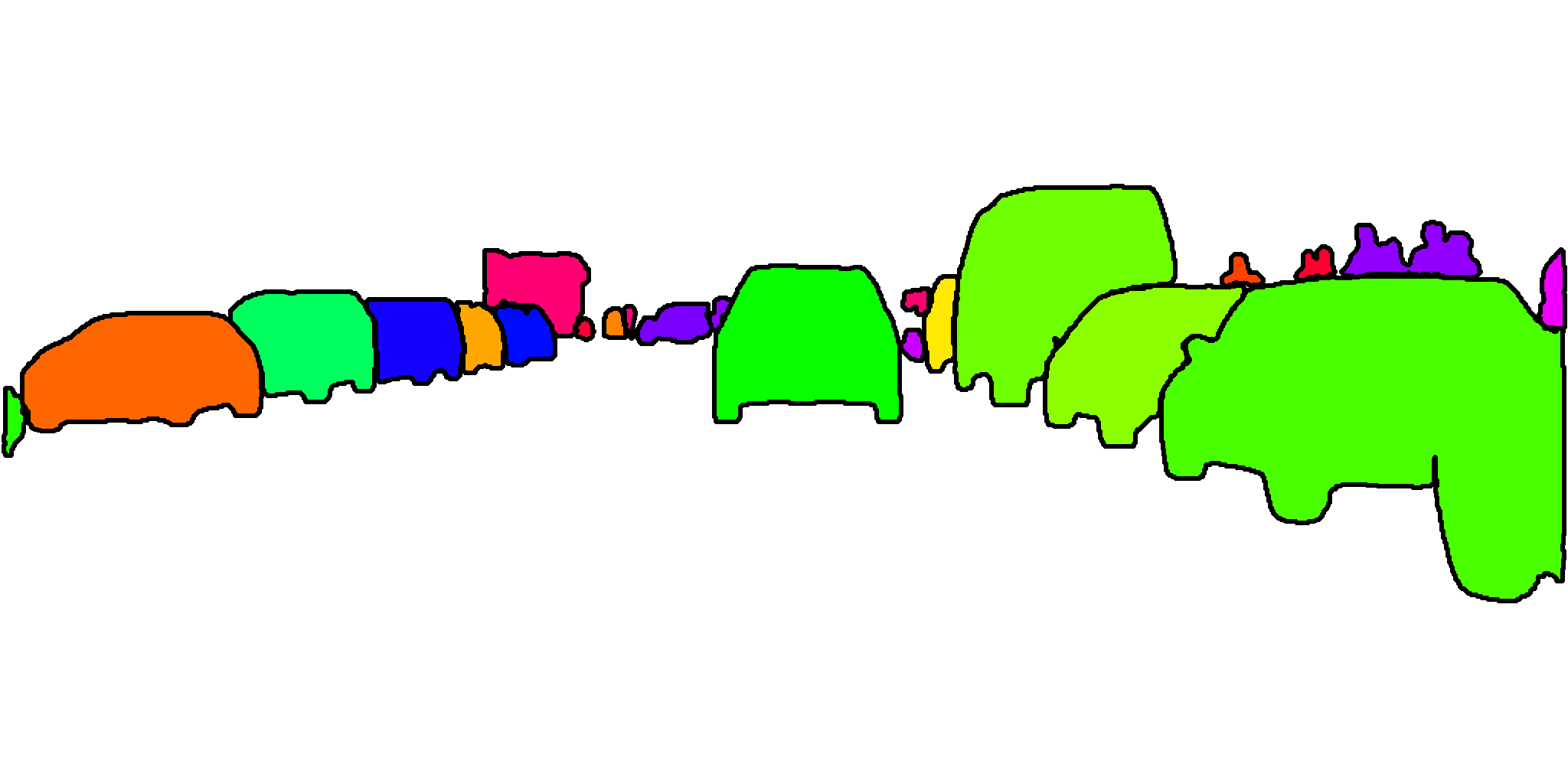}\end{subfigure}
\begin{subfigure}[b]{0.22\textwidth} \centering \includegraphics[trim={0 3cm 0 1.25cm},clip,width=\textwidth]{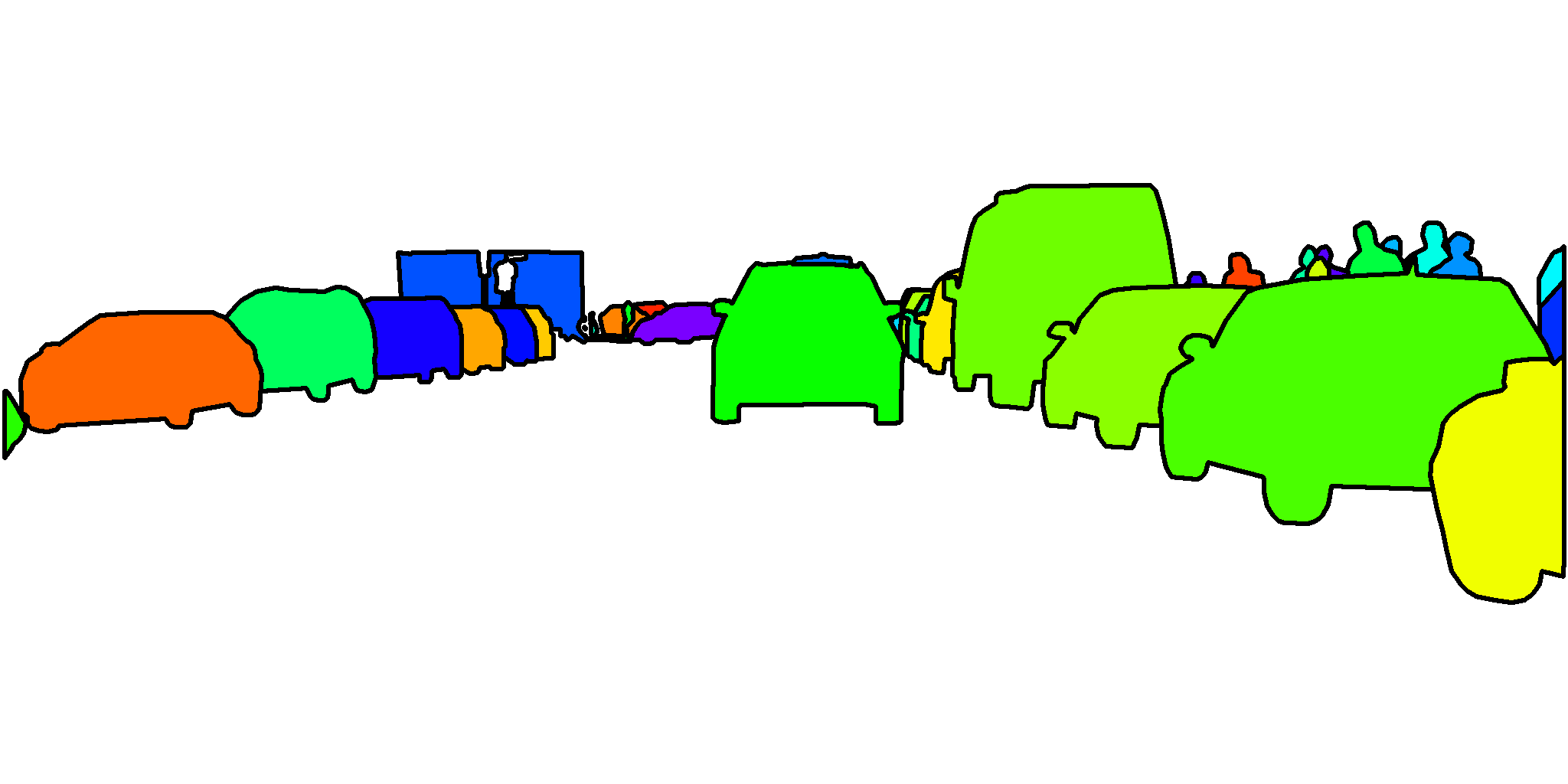}\end{subfigure}

\vspace{0.10cm}

\begin{subfigure}[b]{0.22\textwidth} \centering \includegraphics[trim={0 2.75cm 0 1.25cm},clip,width=\textwidth]{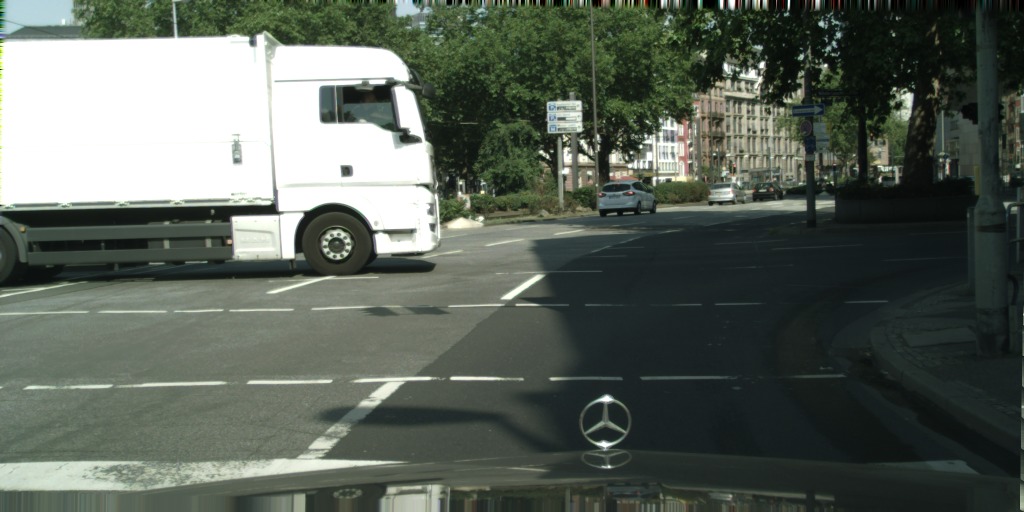}\caption{Input Image}\end{subfigure}
\begin{subfigure}[b]{0.22\textwidth} \centering \includegraphics[trim={0 3cm 0 1.25cm},clip,width=\textwidth]{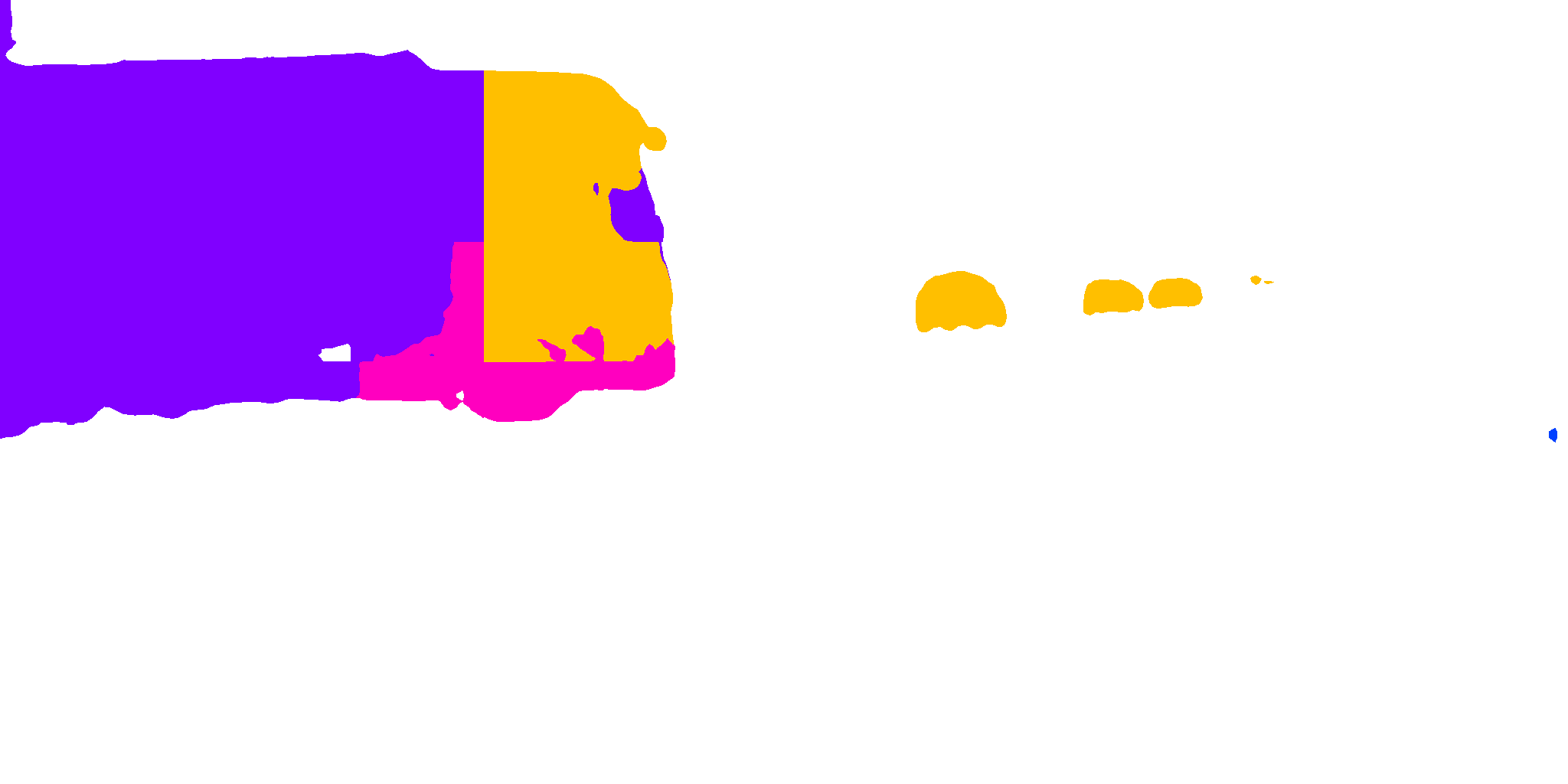}\caption{Sem. Segmentation \cite{PSPNet}}\end{subfigure}
\begin{subfigure}[b]{0.22\textwidth} \centering \includegraphics[trim={0 3cm 0 1.25cm},clip,width=\textwidth]{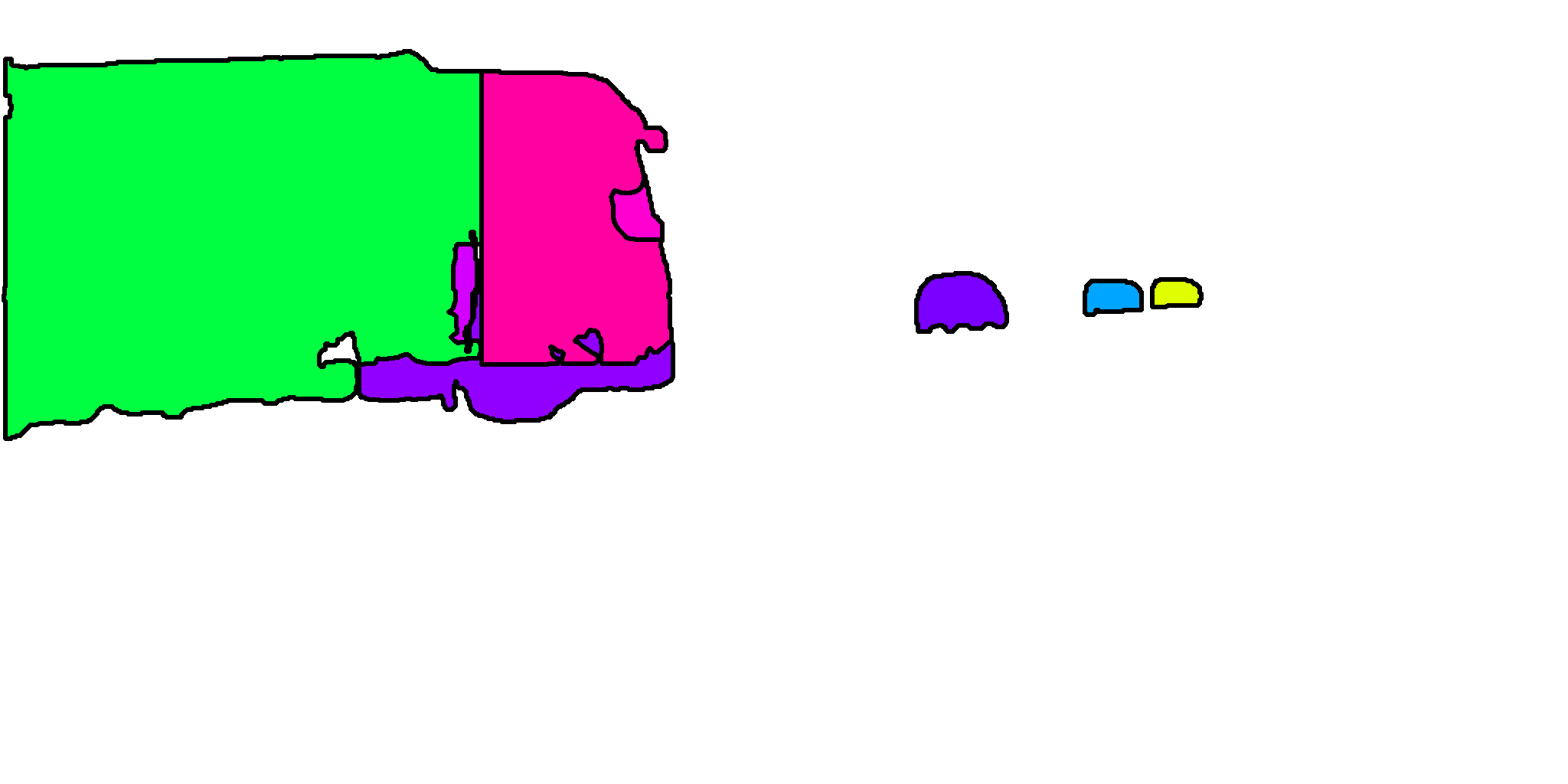}\caption{Our Instance Segmentation}\end{subfigure}
\begin{subfigure}[b]{0.22\textwidth} \centering \includegraphics[trim={0 3cm 0 1.25cm},clip,width=\textwidth]{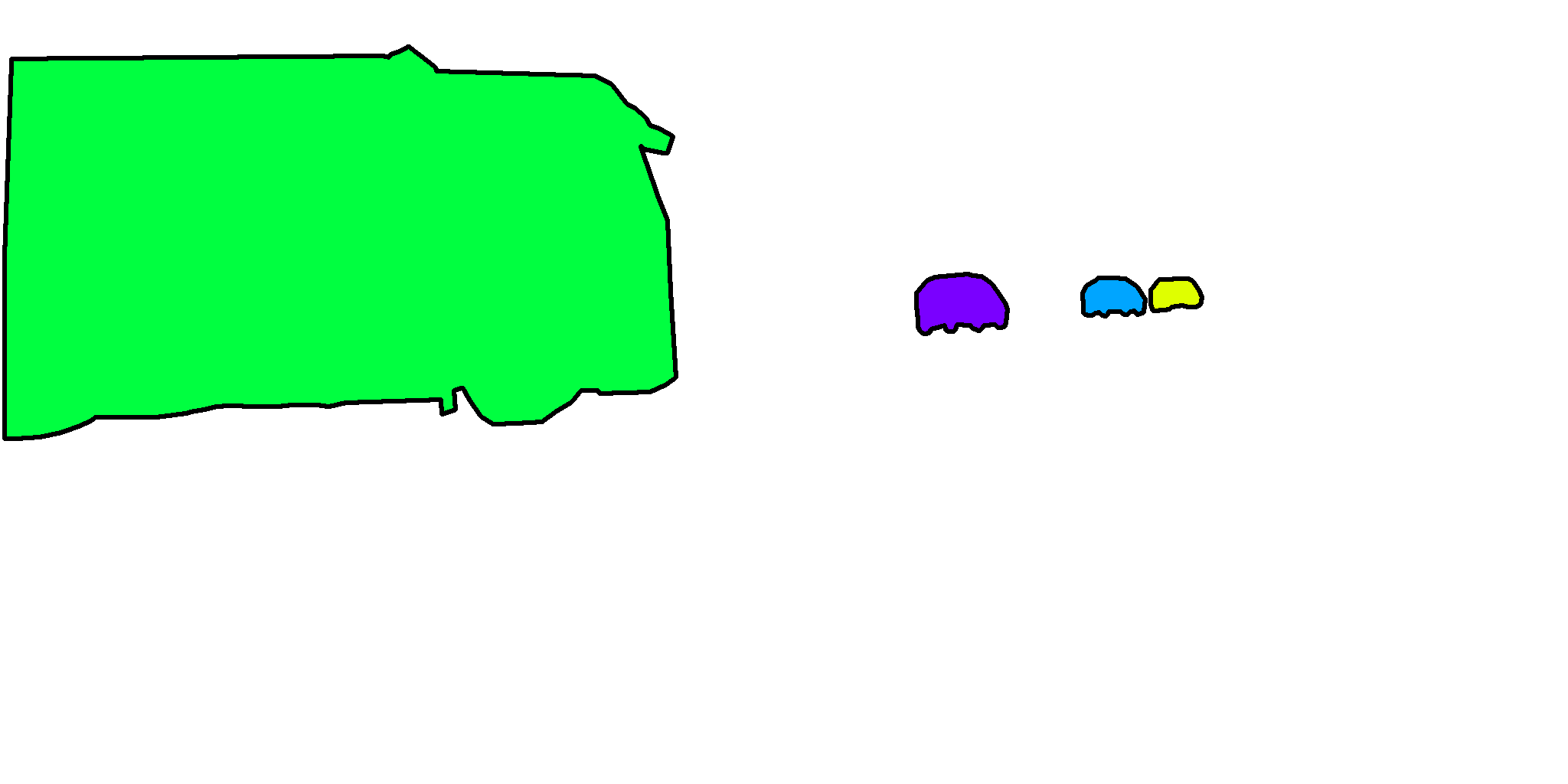}\caption{GT Instance Segmentation}\end{subfigure}
\caption{Sample output of our model on the validation set. Note that predicted object instances and ground truth object instances are only given the same color if they have over 50\% IoU. }\label{fig:qualitative}
\end{center}
\end{figure*}

%% file: conclusion.tex

\section{Conclusion}

In this paper, we have proposed a simple instance segmentation technique inspired by the intuitive and classical watershed transform. Using a novel deep convolutional neural network and innovative loss functions for pre-training and fine-tuning, we proposed a model that generates a modified watershed energy landscape. From this energy landscape, we directly extract high quality object instances. Our experiments show that we can more than double the performance of the state-of-the-art in the challenging Cityscapes Instance Segmentation task. 
We will release the network weights and code for training and testing our model.
In the future, we plan to augment the method to handle object instances bisected by occlusions. Additionally, we wish to explore the possibility of extending our approach to perform joint semantic and instance level segmentation, in hopes of further refining both outputs simultaneously.